\documentclass[format=acmsmall, review=false, screen=true]{acmart}

\usepackage{booktabs} 
\usepackage{graphicx, caption, subcaption}
\usepackage{multirow}
\usepackage{wrapfig}

\usepackage[ruled]{algorithm2e} 

\SetAlFnt{\small}
\SetAlCapFnt{\small}
\SetAlCapNameFnt{\small}
\SetAlCapHSkip{0pt}
\IncMargin{-\parindent}

\setcopyright{acmlicensed}

\acmDOI{0000001.0000001}

\begin{document}
\title[From Selective Deep Convolutional Features to Compact Binary  Representations]{From Selective Deep Convolutional Features\\   to Compact Binary  Representations for Image Retrieval}

\author{Thanh-Toan Do$^*$}
\affiliation{%
  \institution{University of Liverpool}
  \country{United Kingdom}}
\email{thanh-toan.do@liverpool.ac.uk}
\author{Tuan Hoang$^*$}
\affiliation{%
  \institution{Singapore University of Technology and Design}
  \country{Singapore}
}
\email{nguyenanhtuan_hoang@mymail.sutd.edu.sg}

\author{Dang-Khoa Le Tan}
\affiliation{%
 \institution{Singapore University of Technology and Design} 
 \country{Singapore}}
\email{letandang_khoa@sutd.edu.sg}
\author{Huu Le}
\affiliation{%
 \institution{Queensland University of Technology} 
 \country{Australia}}
\email{huu.le@qut.edu.au}
\author{Tam V. Nguyen}
\affiliation{%
 \institution{University of Dayton} 
 \country{United States}}
\email{tamnguyen@udayton.edu}
\author{Ngai-Man Cheung}
\affiliation{%
  \institution{Singapore University of Technology and Design} 
  \country{Singapore}}
\email{ngaiman_cheung@sutd.edu.sg}

\begin{abstract}
In the large-scale image retrieval task, the two most important requirements are the discriminability of image representations and the efficiency in computation and storage of representations. Regarding the former requirement, Convolutional Neural Network (CNN) is proven to be a very powerful tool to extract highly-discriminative local descriptors for effective image search. Additionally, in order to further improve the discriminative power of the descriptors, recent works adopt fine-tuned strategies. In this paper, taking a different approach, we propose a novel, computationally efficient, and competitive framework.
Specifically, we firstly propose various strategies to compute masks, namely \textbf{\textit{SIFT-mask}}, \textbf{\textit{SUM-mask}}, and \textbf{\textit{MAX-mask}}, to select a representative subset of local convolutional features and eliminate redundant features. 
Our in-depth analyses demonstrate that proposed masking schemes are effective to address the burstiness drawback and improve retrieval accuracy. 
Secondly, we propose to employ recent embedding and aggregating methods which can significantly boost the feature discriminability. 
Regarding the computation and storage efficiency, we include a hashing module to produce very compact binary image representations.
Extensive experiments on six image retrieval benchmarks demonstrate that our proposed framework achieves the state-of-the-art retrieval performances. 
\end{abstract}

%
%


%
%


\thanks{$^*$ indicates equal contribution.}

\maketitle

\renewcommand{\shortauthors}{Thanh-Toan Do$^*$, Tuan Hoang$^*$, et al.}

\section{Introduction}
Content-based image retrieval has been an active research field for decades and attracted a sustained attention from the computer vision/multimedia communities 
due to its wide range of applications, e.g., visual search, place recognition. Earlier works heavily rely on hand-crafted local descriptors, e.g., SIFT \cite{SIFT_Lowe} and its variant \cite{rootsift}. Although a lot of great efforts have been made to improve performances of the SIFT-based image search systems, their performances are still limited. There are two main limitations with the SIFT features. The first and the most important one is the low-discriminability of SIFT features \cite{cnn_max_pooling} which is necessary to emphasize the differences in images. Although the limitation have been relieved to some extent by embedding local features to a much higher dimensional space \cite{bow,FisherVector,vlad,Temb,F-FAemb}, the semantic gap between human understanding on objects/scenes and SIFT-based image representation is still considerable large \cite{cnn_max_pooling}. Secondly, the {\em burstiness} effect \cite{burstiness}, i.e., numerous descriptors are almost identical within an image, significantly degrades the quality of SIFT-based image representation \cite{vlad, burstiness,revisitvlad}.

Recently, deep Convolutional Neural Networks (CNN) achieve great success in various problems including image classification \cite{Alexnet,VGG,googlenet,resnet}, semantic segmentation~\cite{journals/corr/LinSRH15,DBLP:conf/iccv/HeGDG17}, object detection \cite{RCNN,FasterRCNN} and image retrieval \cite{cnn_max_pooling,CroW,R-MAC,netvlad,MOF,SCDA}. 
While the output of the deeper layers, e.g., fully-connected, can be helpful for the image retrieval task \cite{MOP}. Recent works \cite{cnn_max_pooling,CroW,R-MAC,netvlad,MOF,SCDA} show that using the outputs of middle layers, e.g., convolution layers, can help to enhance the retrieval performances by larger margins. 

Even though the local convolutional (conv.) features are more discriminative than SIFT features \cite{cnn_max_pooling}, the burstiness issue, which may appear in the local conv. features, has not been investigated previously.
In this paper, by delving deeper into the burstiness issue, we propose three different masking schemes to select \textit{highly-representative} local conv. features and robustly eliminate redundant local features. The masking schemes are named as \textbf{\textit{SIFT-mask}}, \textbf{\textit{SUM-mask}}, and \textbf{\textit{MAX-mask}}. The elimination of redundant local features results in more discriminative representation and efficient computation, we will further discuss these advantages in the experiment section.
The fundamentals of our proposal are that we utilize SIFT detector \cite{SIFT_Lowe} to produce SIFT-mask; additionally, we apply sum-pooling and max-pooling across all conv. feature maps to derive SUM-mask and MAX-mask, respectively. Note that our idea of using SIFT coordinate for CNN based image retrieval is novel. Our SUM-mask is also new. Previous works apply sum-pooling within a feature map; our mask is computed by summing across feature maps. Moreover, while max-pooling \cite{R-MAC} gets the maximum value, our MAX-mask obtains the location of that value.

In addition, the majority of recent works, that work on local conv. features \cite{R-MAC,CroW,finetune_hard_samples}, do not utilize feature embedding and aggregating methods \cite{FisherVector,vlad,Temb,F-FAemb}, which are useful steps to boost the discriminability for SIFT features. In \cite{cnn_max_pooling}, the authors discussed that the deep conv. features are already discriminative enough for image retrieval task; hence, the embedding step is unnecessary. 
However, we find that applying the state-of-the-art embedding and aggregating \cite{Temb,F-FAemb,vlad,FisherVector} can significantly help to enhance the discriminability of image representations. 
Therefore, by applying embedding and aggregating on our selective local conv. features, the aggregated representations improve image retrieval performance significantly.


\begin{figure*}[t]
\centering
\includegraphics[width=\textwidth]{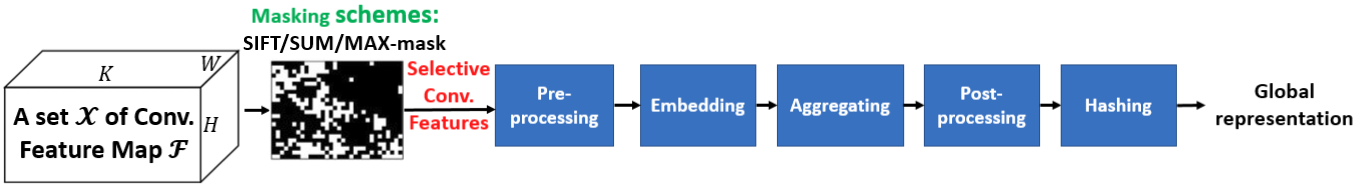}
\caption{The overview of our proposed framework to produce discriminative global binary representations.}
\label{fig:framework}
\vspace{-1.5em}
\end{figure*}

Furthermore, in order to achieve compact binary codes, we cascade a state-of-the-art unsupervised hashing method, e.g., Iterative Quantization (ITQ) \cite{ITQ}, Relaxed Binary Autoencoder (RBA) \cite{SAH}, Simultaneous Compression and Quantization (SCQ) \cite{SCQ}, into the proposed system. 
The binary representations would help to achieve significant benefits in retrieval speed and memory consumption. Fig.~\ref{fig:framework} presents the overview of the proposed framework. In summary, in this work we make following contributions. 

\textbf{Contributions. } {A preliminary version of this work  has been presented in~\cite{DBLP:journals/corr/HoangDTC17aa}. 
In the preliminary version \cite{DBLP:journals/corr/HoangDTC17aa}, we first propose various novel masking schemes, which are proven to effectively eliminate redundant local conv. features. Secondly, we leverage the state-of-the-art embedding and aggregating methods to produce highly-discriminative global image representations. We comprehensively evaluate different components to build an efficient framework that achieves the state-of-the-art retrieval performance on standard benchmark datasets when using real-value image representations.}
{In this current version, we introduce additional contributions as follows: Firstly, we conduct analysis to explain how various masking schemes work, both qualitatively and quantitatively. Secondly, we show that assembling information of different abstract levels is beneficial in the image retrieval task as this could help to produce more informative and discriminative representations. Thirdly, we then further optimize the framework to solve two crucial problems for large scale image search, i.e., searching speed and storage. 
Specifically, we propose to cascade a state-of-the-art unsupervised hash function into the framework to further binarize real-valued aggregated representations to binary representations. Note that binary representations would allow the fast searching and efficient storage which are very critical in large scale search systems. In addition, we also conduct very large scale experiments, i.e., on Flickr1M dataset~\cite{herve_eccv2008}, which consists of over one million images. 
The experiments on such kind of large scale dataset are necessary to confirm the effectiveness of the proposed method for real applications which usually have to deal with very large scale datasets. By the best of our knowledge, our work is the first deep learning-based retrieval method which conducts the evaluation on that kind of large scale dataset. 
We also conduct more experiments to deeply analyze the effectiveness of the proposed framework and to extensively compare to the state of the art.   
The extensive experiments on six benchmark datasets show that the proposed framework significantly outperforms the state of the art when images are represented by either real-valued representations or compact binary representations.}

We organize the remainders of this paper as follows. Section \ref{sec:related-work} presents related works. Section \ref{sec:selective_local_fea} presents our main contributions of the proposed masking schemes. Section \ref{sec:frame} presents the proposed framework for computing the final image representation from a set of selected local deep conv. features. Section \ref{sec:exp} presents comprehensive experiments to evaluate our proposed framework.
Section \ref{sec:conclusion} concludes the paper.

\section{Related work}
\label{sec:related-work}


In the last few years, image retrieval has witnessed an increasing of performance due
to the use of better image representations, i.e., deep features obtained from pre-trained CNN models, which are trained on image classification task. The early CNN-based work  \cite{DBLP:conf/cvpr/RazavianASC14} directly used deep fully-connected (FC) activations for the image retrieval. 
Instead of directly using features from the pre-trained networks for the retrieval as~\cite{DBLP:conf/cvpr/RazavianASC14}, other works apply different processings on the pre-trained features to enhance the discriminability. Gong \textit{et al.} \cite{MOP} proposed Multi-Scale Orderless Pooling to embed and aggregate CNN FC activations of image patches of an image at different scales. Hence, the final features are more invariant to the scale.
However, as multiple patches (cropped and resized to a specific size) of an image are fed forward into the CNN, the method endures a high computational cost. Yan \textit{et al.} \cite{CNN-vs-SIFT} revisted the SIFT feature and suggested that SIFT and CNN FC features are highly complementary. Therefore, they proposed to integrate SIFT features with CNN FC features at multiple scales. Concurrently, Liu \textit{et al.} \cite{imageGraph} proposed ImageGraph to fuse various types of features, e.g., CNN FC features, BoW on SIFT \cite{SIFT_Lowe} descriptors, HSV color histogram,  and GIST features \cite{gist}. This method even though achieves very good performances, it requires very high-dimensional features. Furthermore, ImageGraph must be built on database images, which may be prohibitive on large scale datasets.

Recently, many image retrieval works shift the attention from FC features to conv. features. This is because outputs of lower layers contain more general information and spatial information is still preserved \cite{generic2specific}. 
In this case, the conv. features are considered as local features. Hence, the sum-pooling or max-pooling method is usually applied 
to achieve a single representation.
Babenko and Lempitsky \cite{cnn_max_pooling} demonstated that by whitening the final image representation, sum-pooling can outperform max-pooling.
Kalantidis \textit{et al.} \cite{CroW} proposed to learn weights for both feature channels and spatial locations which helps to enhance the discriminability of sum-pooling representation on conv. features.
Tolias \textit{et al.} \cite{R-MAC} revisited max-pooling by proposing the strategy to aggregate the maximum activation over multiple spatial regions sampled on a output of a conv. layer using a fixed layout. 
Similarly, Jian Xu \textit{et al.} \cite{PWA} proposed to aggregate features which are weighted using probabilistic proposals.

Instead of using pre-trained features (with / without additional processing) for the retrieval task. In \cite{neuralcode}, Babenko \textit{et al.} showed that fine-tuning an off-the-shelf network (e.g., AlexNet \cite{Alexnet} or VGG \cite{VGG}) can produce more discriminative deep features \cite{neuralcode} for the image search task. However, collecting labeled training data is non-trivial \cite{neuralcode}. Recent works tried to overcome this challenge by proposing unsupervised/weakly-supervised fine-tuning approaches which are specific for image retrieval.  
Arandjelovic \textit{et al.} \cite{netvlad} proposed the NetVLAD architecture which can be trained in an end-to-end fashion. The author also proposed to collect from Google Street View Time Machine in a weakly-supervised process. 
Adopting a similar approach,
Cao \textit{et al.} \cite{quartet-net} proposed to harvest data from Flickr with GPS information to form GeoPair dataset \cite{goepair}. The dataset is afterward used to train the special Quartet-net architecture.
Radenovic \textit{et al.} \cite{finetune_hard_samples}, concurrently, proposed a different approach to fine-tune a pretrained CNN on classification task for image retrieval. The authors propose to use 3D reconstruction to obtain matching / non-matching image pairs in an unsupervised manner for fine-tuning process. 
Recently, Noh \textit{et al.} \cite{DELF} proposed the DEep Local Features (DELF) pipeline with attention-based keypoint selection for large scale image retrieval. The model is fine-tuned using their proposed Google Landmark dataset. However, the pipeline requires the geometric verification using RANSAC. Even though,  features are compressed to very low dimensions, e.g., 40-D, for the trade-off between compactness, speed and discrimination, the process is still computation and memory intensive. 
{Besides, the self-supervised approach \cite{8310624,norooziECCV16} to fine-tuning the models is also an interesting approach to enhance the discrimination power of the CNN models for the image retrieval task.}

In regards to compact image representations, the earlier work \cite{FeaDimSelect} presented feature dimension selection on embedded high-dimensional features as a compression method to achieve compact representations. Radenovic \textit{et al.} \cite{finetune_hard_samples,ft_nohuman} later introduced to learn the whitening and dimensionality reduction in the supervised manner resulting in better performances than the baseline PCA method. Albert \textit{et al.} \cite{Gordo2017} made use of the product quantization \cite{PQ} to compress image representations. This approach even though achieves good accuracy, it is not as efficient as the hashing approach, which we utilize in this paper, in term of retrieval time \cite{PolysemousCodes}. Do \textit{et al.} \cite{SAH} proposed to produce binary representations by simultaneously aggregating raw local features and hashing. Differentially, in this paper, we proposed various masking schemes in combination with a complete framework to produce more discriminative binary representations.
Taking similar approach with \cite{finetune_hard_samples} to mining the training datasets of matching / non-matching image pairs, Do \textit{et. al.}  \cite{P2B} proposed to directly learn the compact binary codes from input images.
{In BGAN \cite{BGAN}, the authors utilize Generative Adversarial Networks (GAN) \cite{GAN} to generate binary codes that can well represented for images in the retrieval task. Recently, Song \textit{et al.} proposed Deep Region Hashing (DRH) \cite{DRH} which computes binary codes for both global and local features. In which the global binary codes are used to obtain initial ranking, and the local binary codes are used for regional re-search (re-ranking). Similar to \cite{R-MAC}, re-ranking can help to improve performance; however, this approach results in significant increases in storage as all local binary codes are requires to be stored. Additionally, additional processing time is also required. Hence, this approach may not be scalable for very large-scale databases, e.g., millions or billions of images.}
{In addition to the quantization and hashing methods, in Quantization-Based Hashing (QBH) \cite{SONG2018175}, the author proposed a novel approach to combine the advantages of quantizaton-based methods and hashing methods.
We would like to refer readers to \cite{sift_cnn} for a more comprehensive survey on image retrieval.}
\section{Selective Local Deep Convolution Features}
\label{sec:selective_local_fea}

\def \imagescale {0.19}
\def \imageheight {2.cm}
\begin{figure*}[p]
\centering

(1) \begin{subfigure}[b]{\imagescale\textwidth}
\caption{}
\frame{\includegraphics[width=\textwidth]{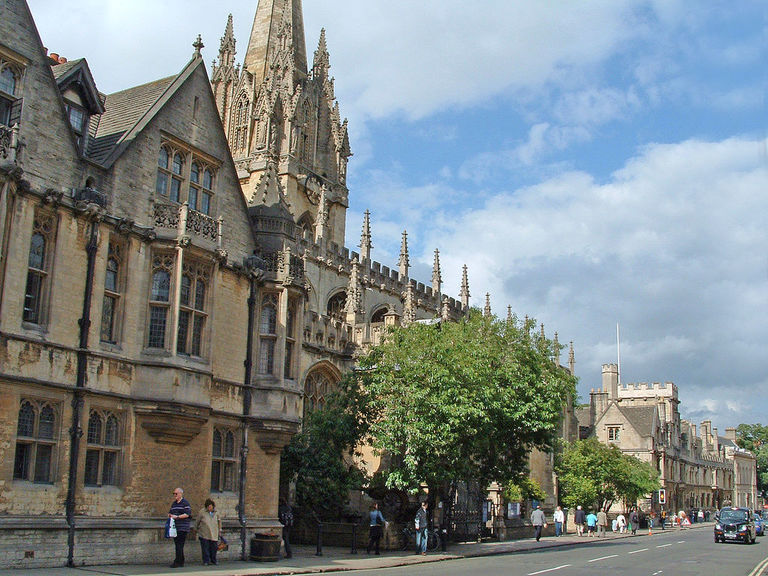}}
\end{subfigure}
\qquad
\begin{subfigure}[b]{\imagescale\textwidth}
\caption{}
\frame{\includegraphics[height=\imageheight,width=\textwidth]{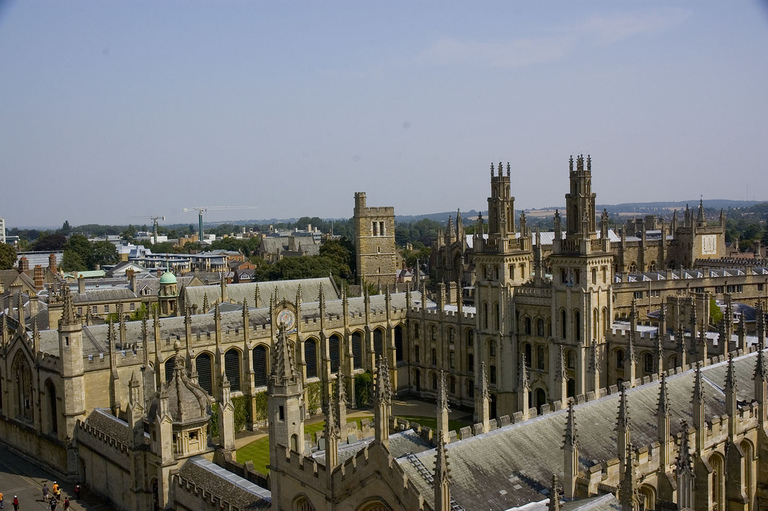}}
\end{subfigure}
\qquad
\begin{subfigure}[b]{\imagescale\textwidth}
\caption{}
\frame{\includegraphics[width=\textwidth]{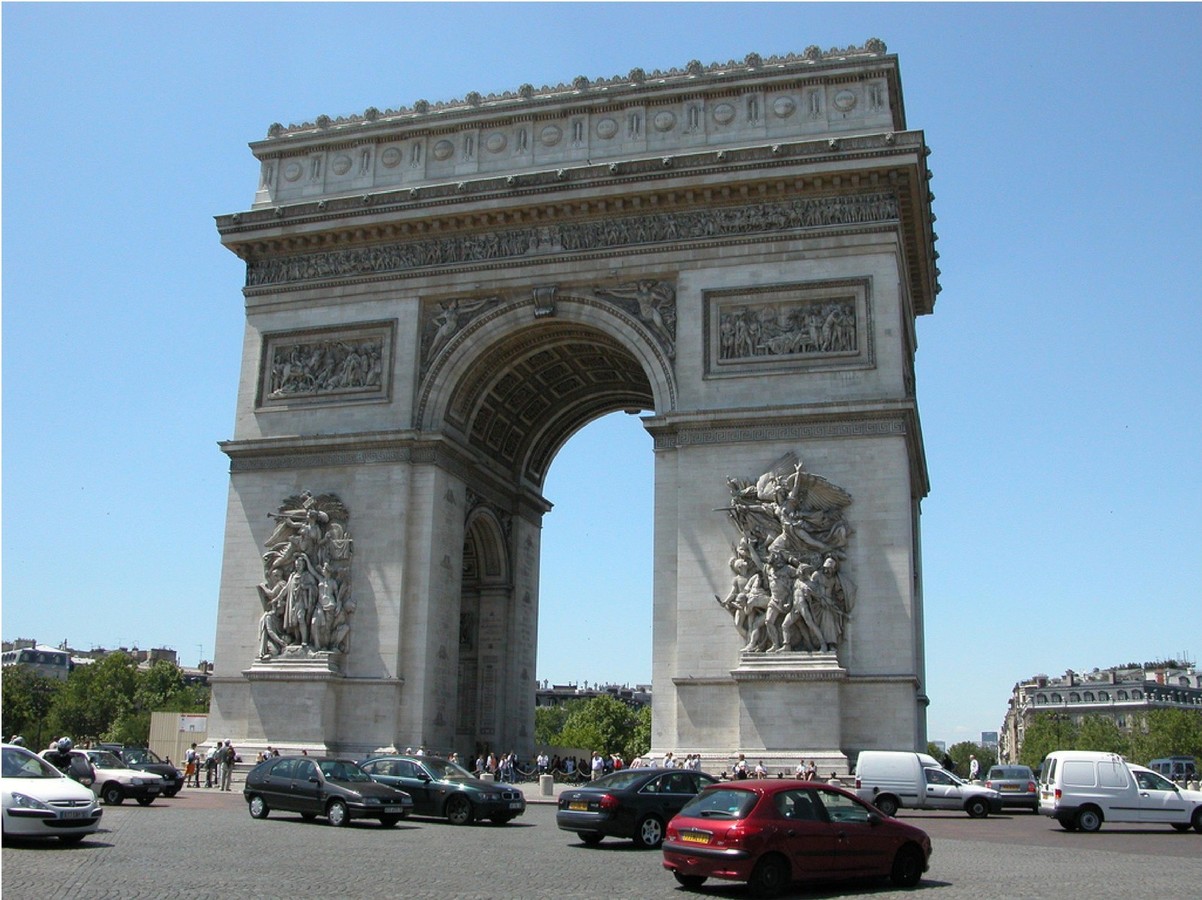}}
\end{subfigure}
\qquad
\begin{subfigure}[b]{\imagescale\textwidth}
\caption{}
\frame{\includegraphics[width=\textwidth]{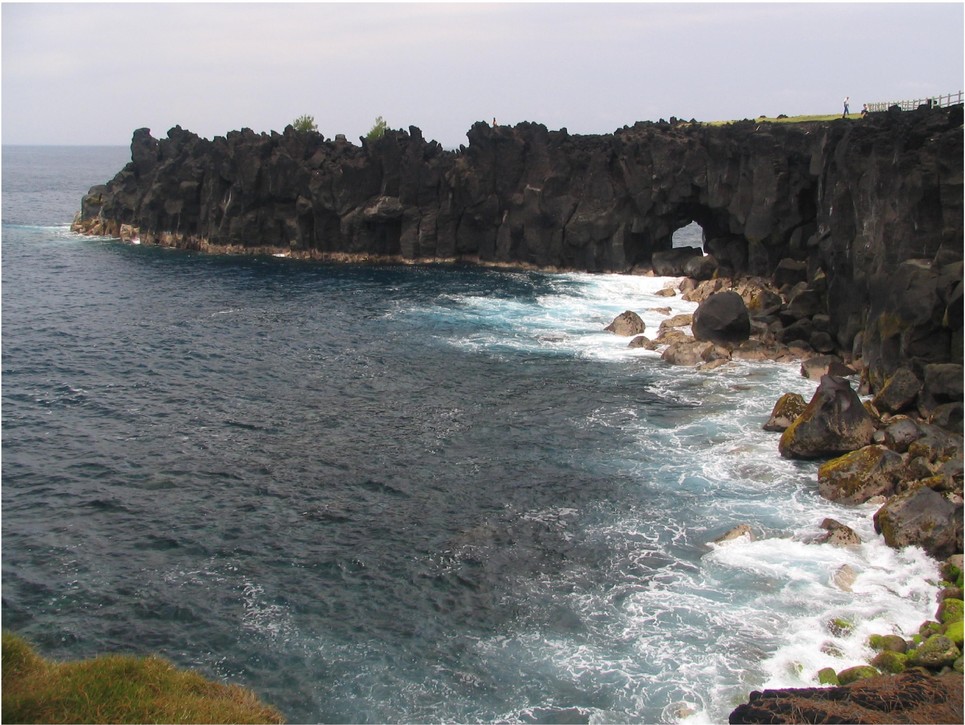}}
\end{subfigure}

\smallskip
(2) \begin{subfigure}[b]{\imagescale\textwidth}
\frame{\includegraphics[width=\textwidth]{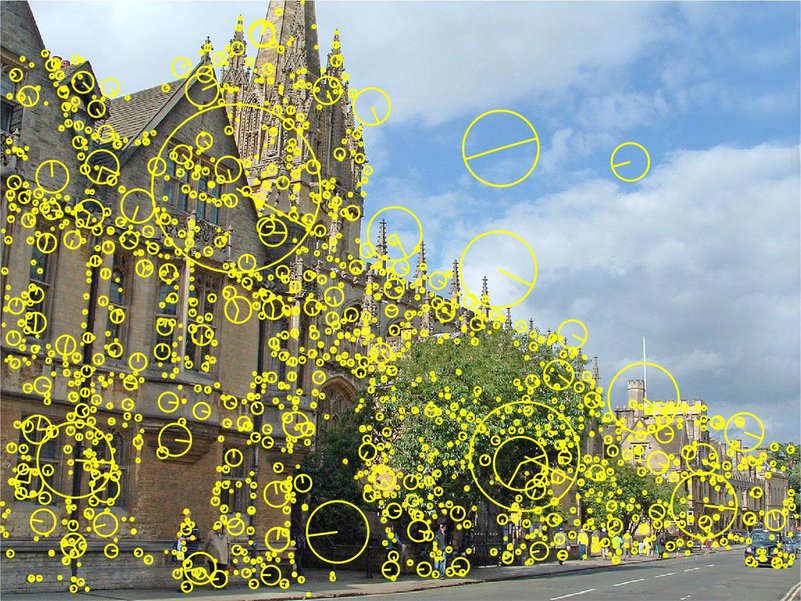}}
\end{subfigure}
\qquad
\begin{subfigure}[b]{\imagescale\textwidth}
\frame{\includegraphics[height=\imageheight,width=\textwidth]{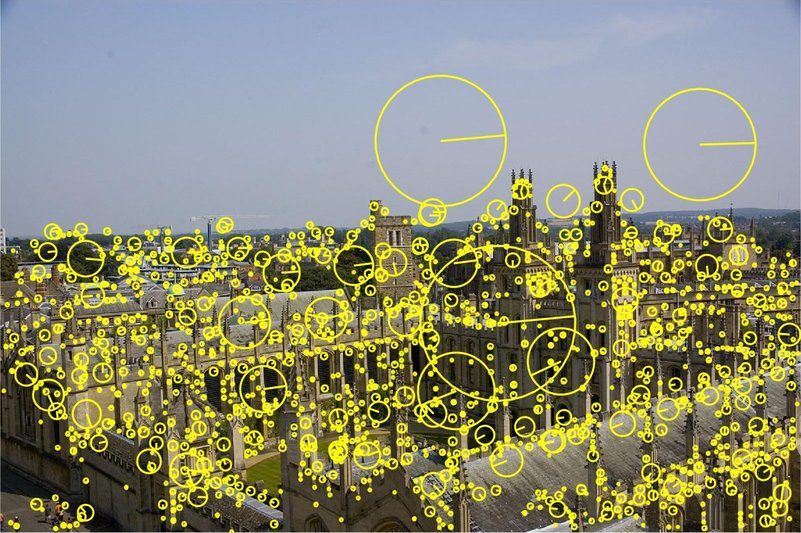}}
\end{subfigure}
\qquad
\begin{subfigure}[b]{\imagescale\textwidth}
\frame{\includegraphics[width=\textwidth]{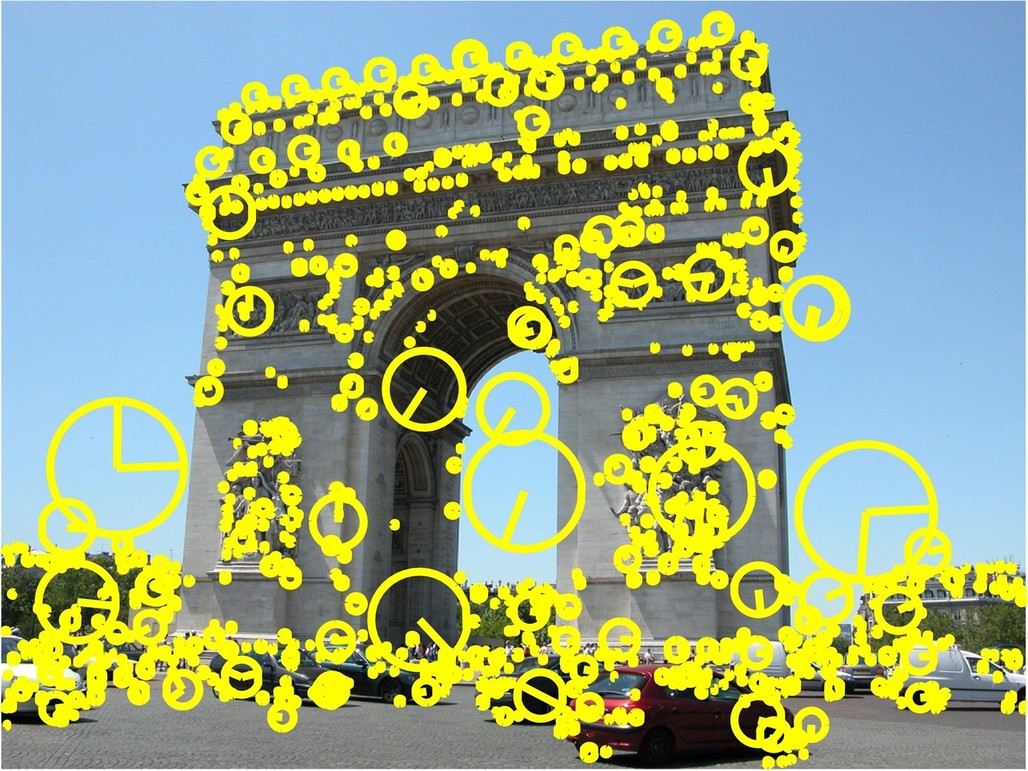}}
\end{subfigure}
\qquad
\begin{subfigure}[b]{\imagescale\textwidth}
\frame{\includegraphics[width=\textwidth]{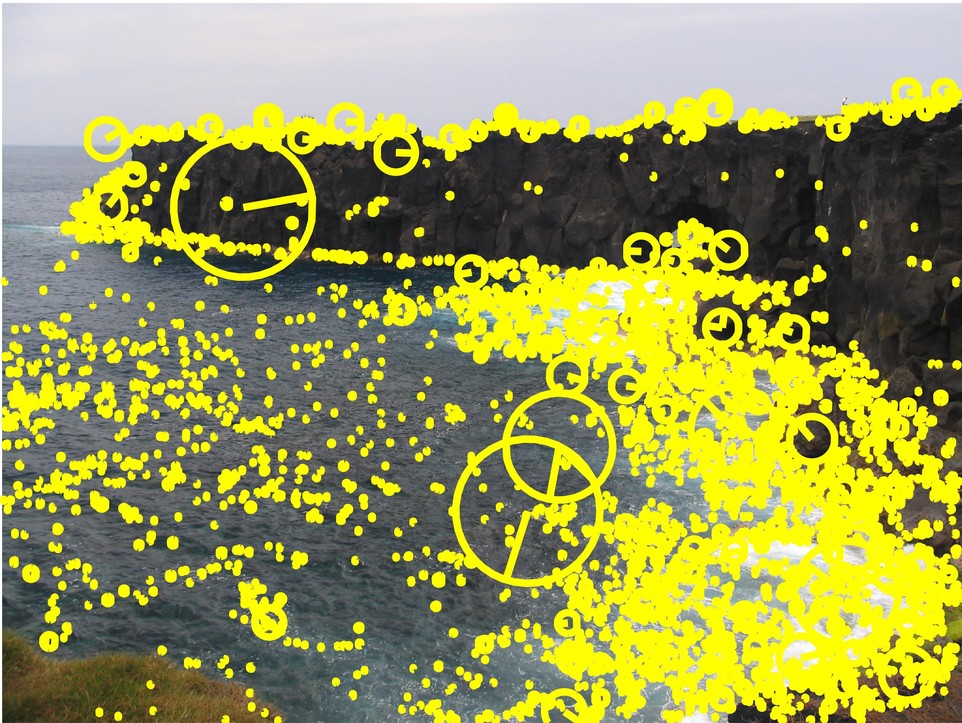}}
\end{subfigure}

\smallskip
(3) \begin{subfigure}[b]{\imagescale\textwidth}
\frame{\includegraphics[width=\textwidth]{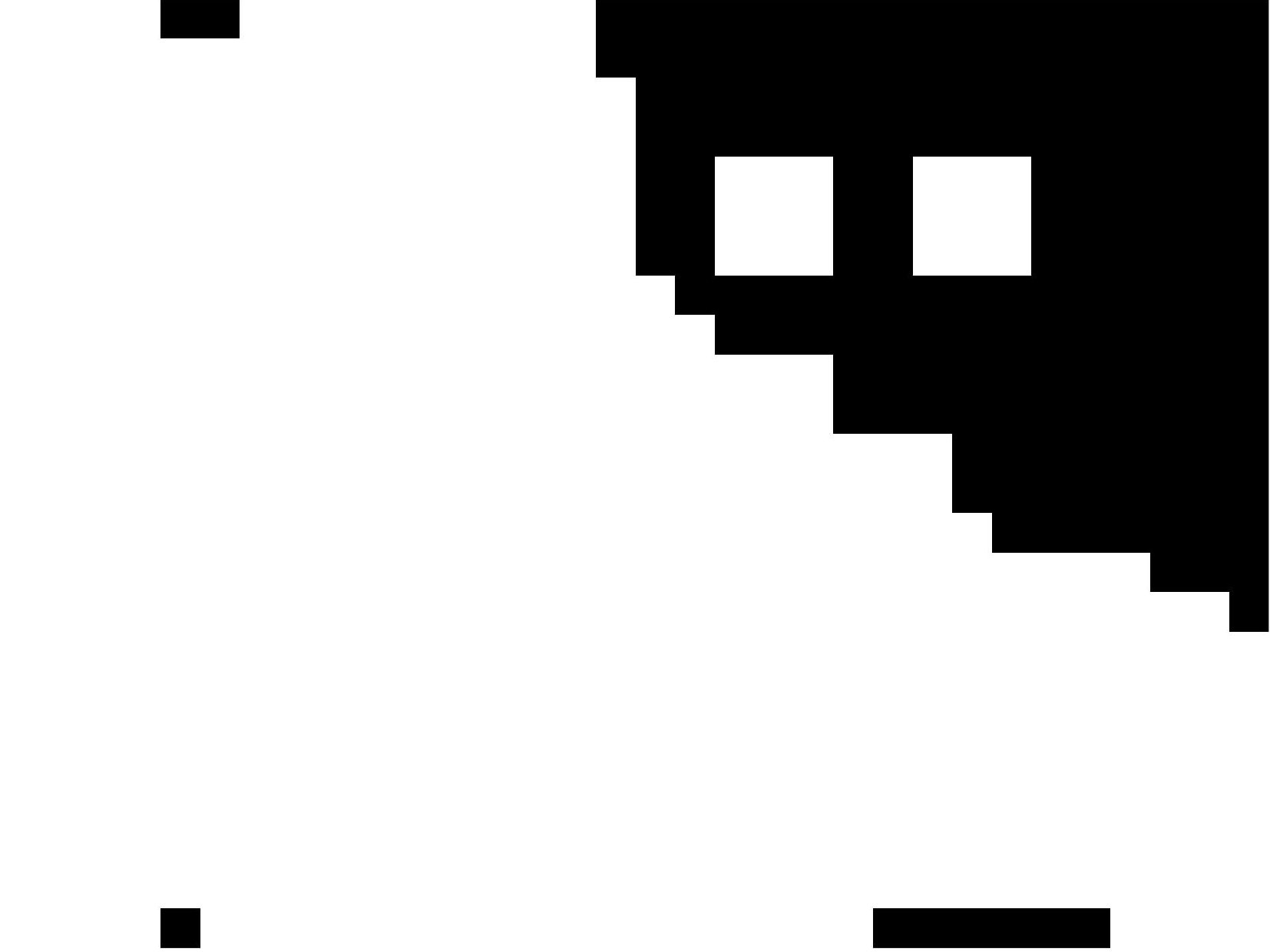}}
\end{subfigure}
\qquad
\begin{subfigure}[b]{\imagescale\textwidth}
\frame{\includegraphics[height=\imageheight,width=\textwidth]{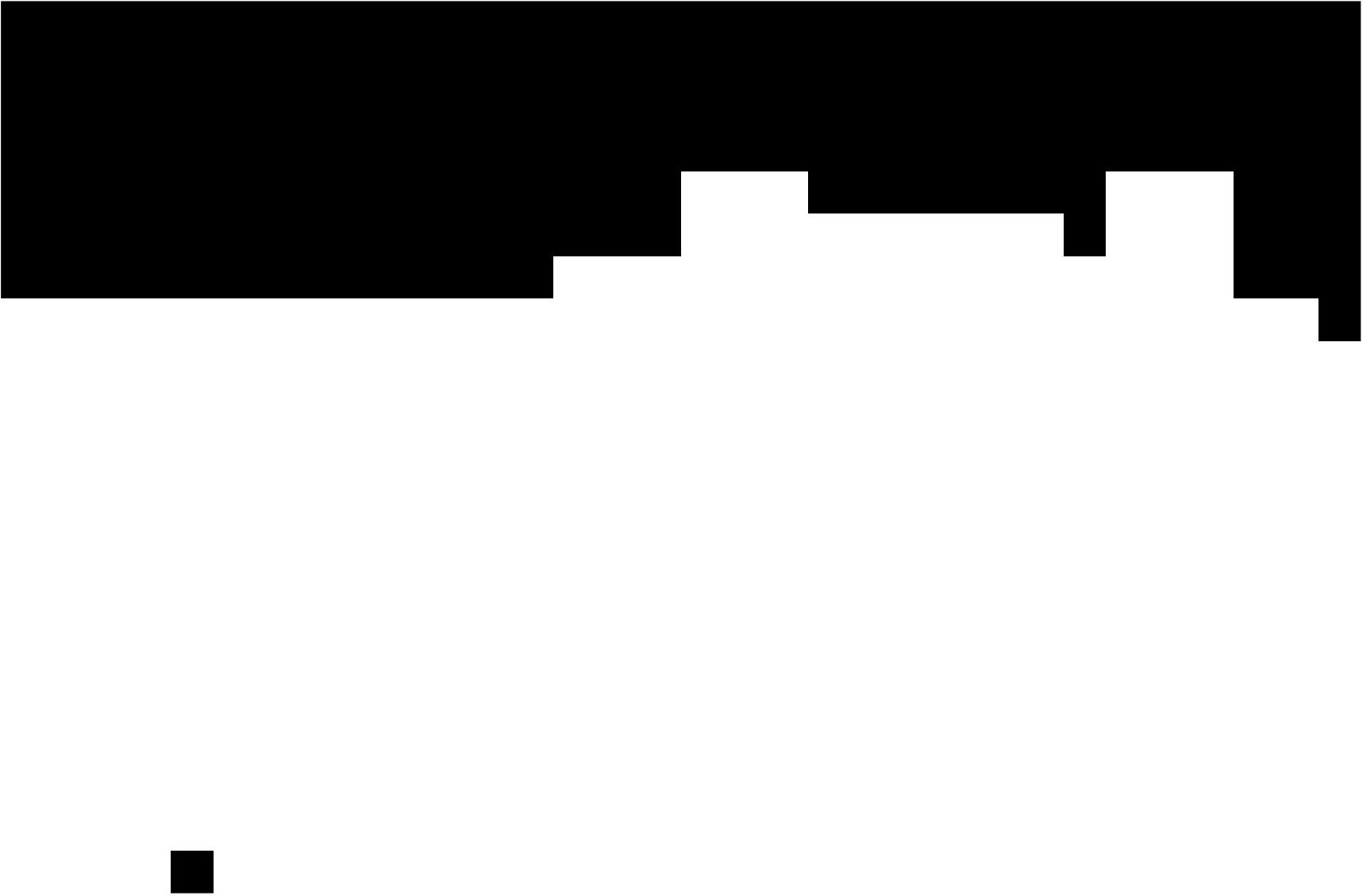}}
\end{subfigure}
\qquad
\begin{subfigure}[b]{\imagescale\textwidth}
\frame{\includegraphics[width=\textwidth]{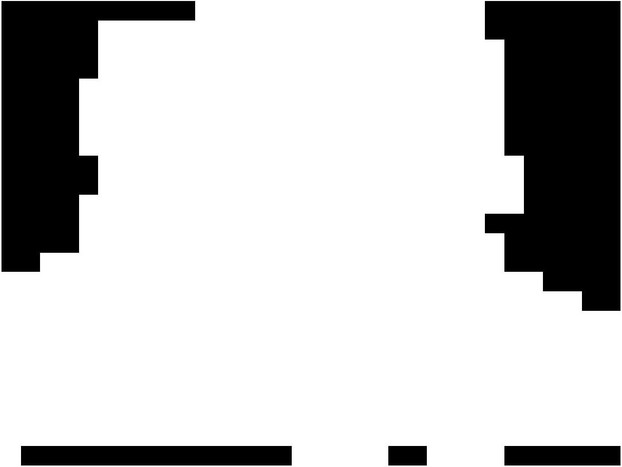}}
\end{subfigure}
\qquad
\begin{subfigure}[b]{\imagescale\textwidth}
\frame{\includegraphics[width=\textwidth]{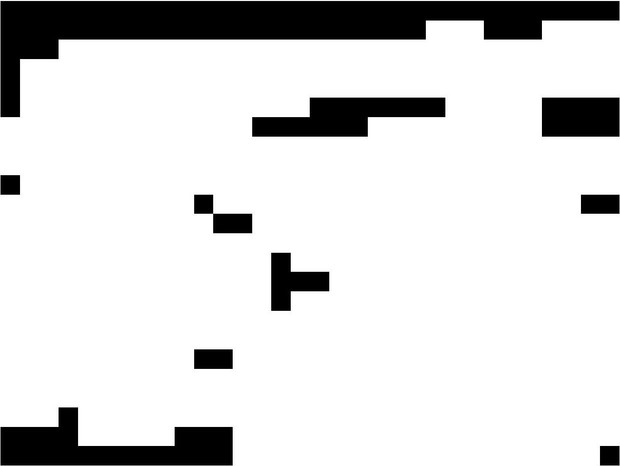}}
\end{subfigure}

\smallskip
(4) \begin{subfigure}[b]{\imagescale\textwidth}
{\includegraphics[width=\textwidth]{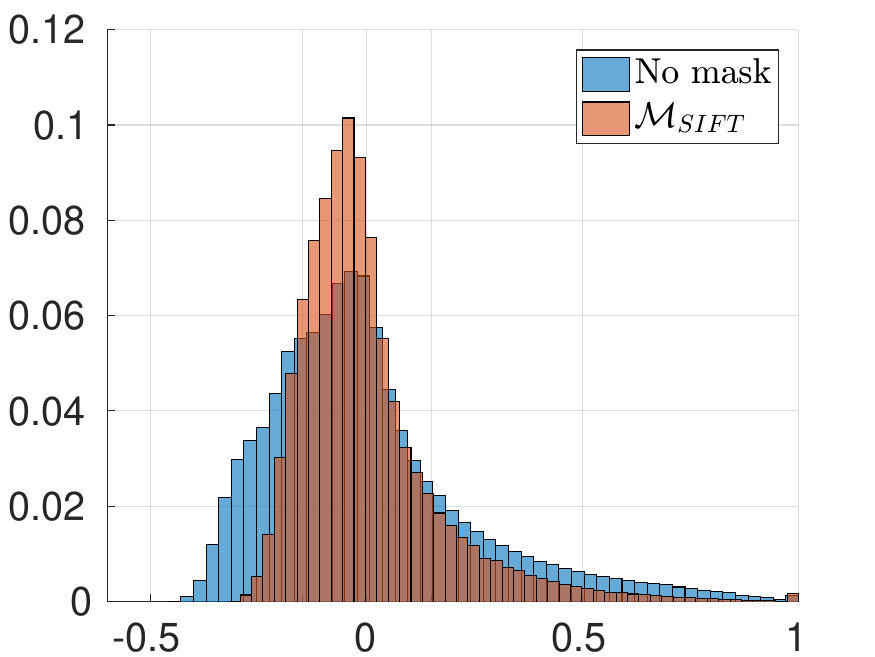}}
\end{subfigure}
\qquad 
\begin{subfigure}[b]{\imagescale\textwidth}
{\includegraphics[width=\textwidth]{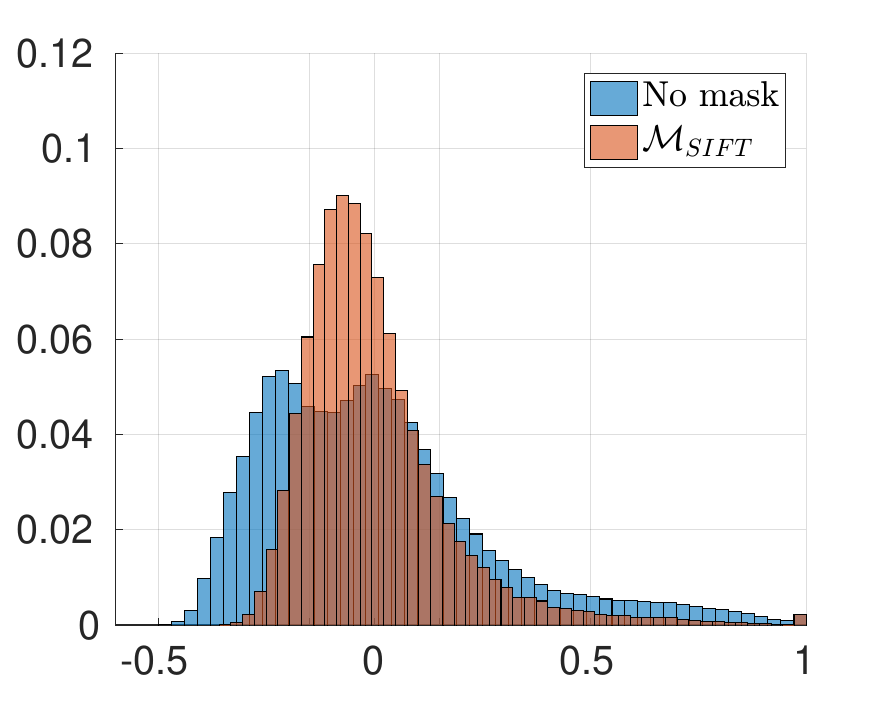}}
\end{subfigure}
\qquad
\begin{subfigure}[b]{\imagescale\textwidth}
{\includegraphics[width=\textwidth]{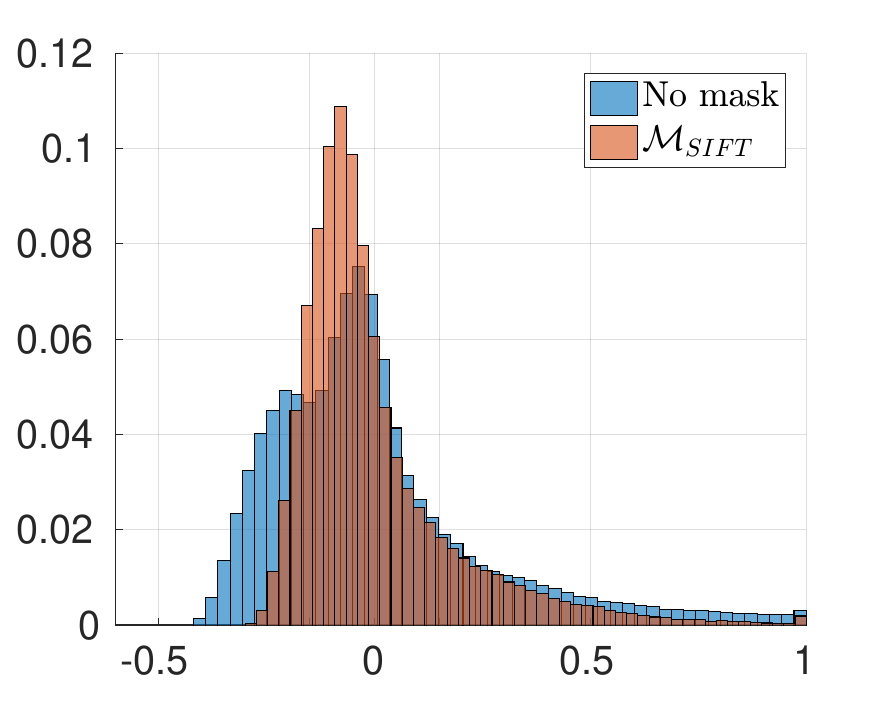}}
\end{subfigure}
\qquad
\begin{subfigure}[b]{\imagescale\textwidth}
{\includegraphics[width=\textwidth]{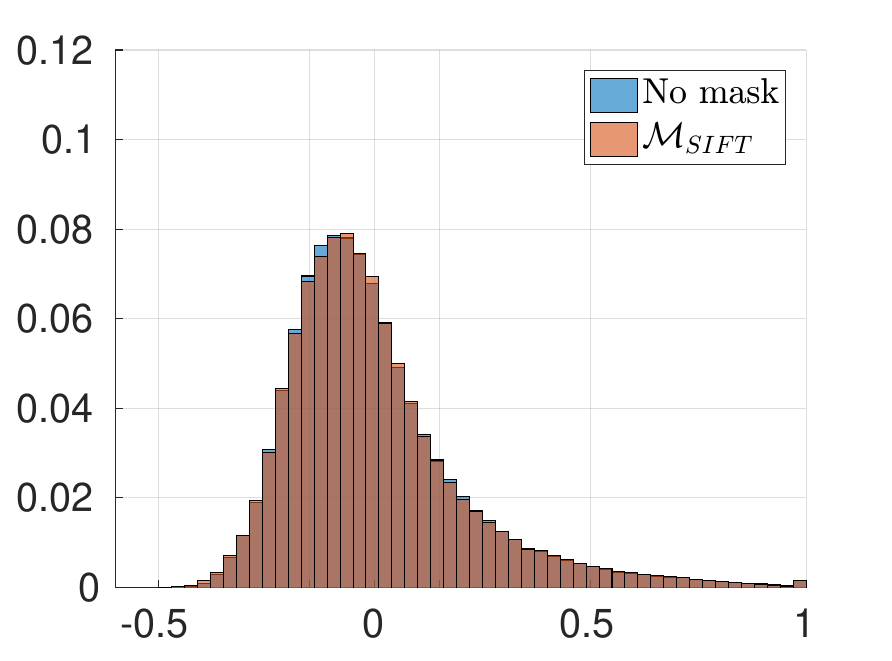}}
\end{subfigure}

\smallskip
(5) \begin{subfigure}[b]{\imagescale\textwidth}
\frame{\includegraphics[width=\textwidth]{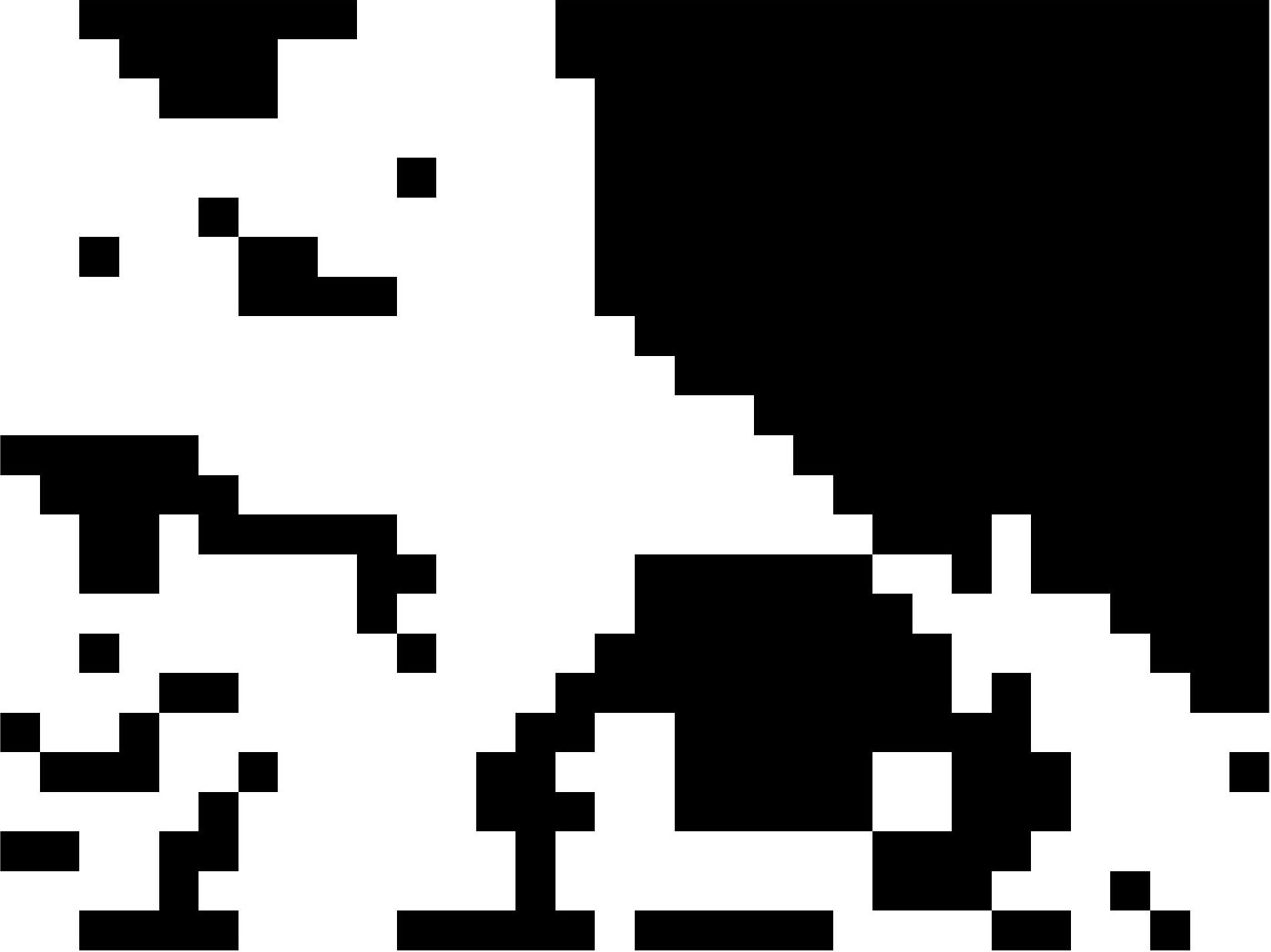}}
\end{subfigure}
\qquad
\begin{subfigure}[b]{\imagescale\textwidth}
\frame{\includegraphics[height=\imageheight,width=\textwidth]{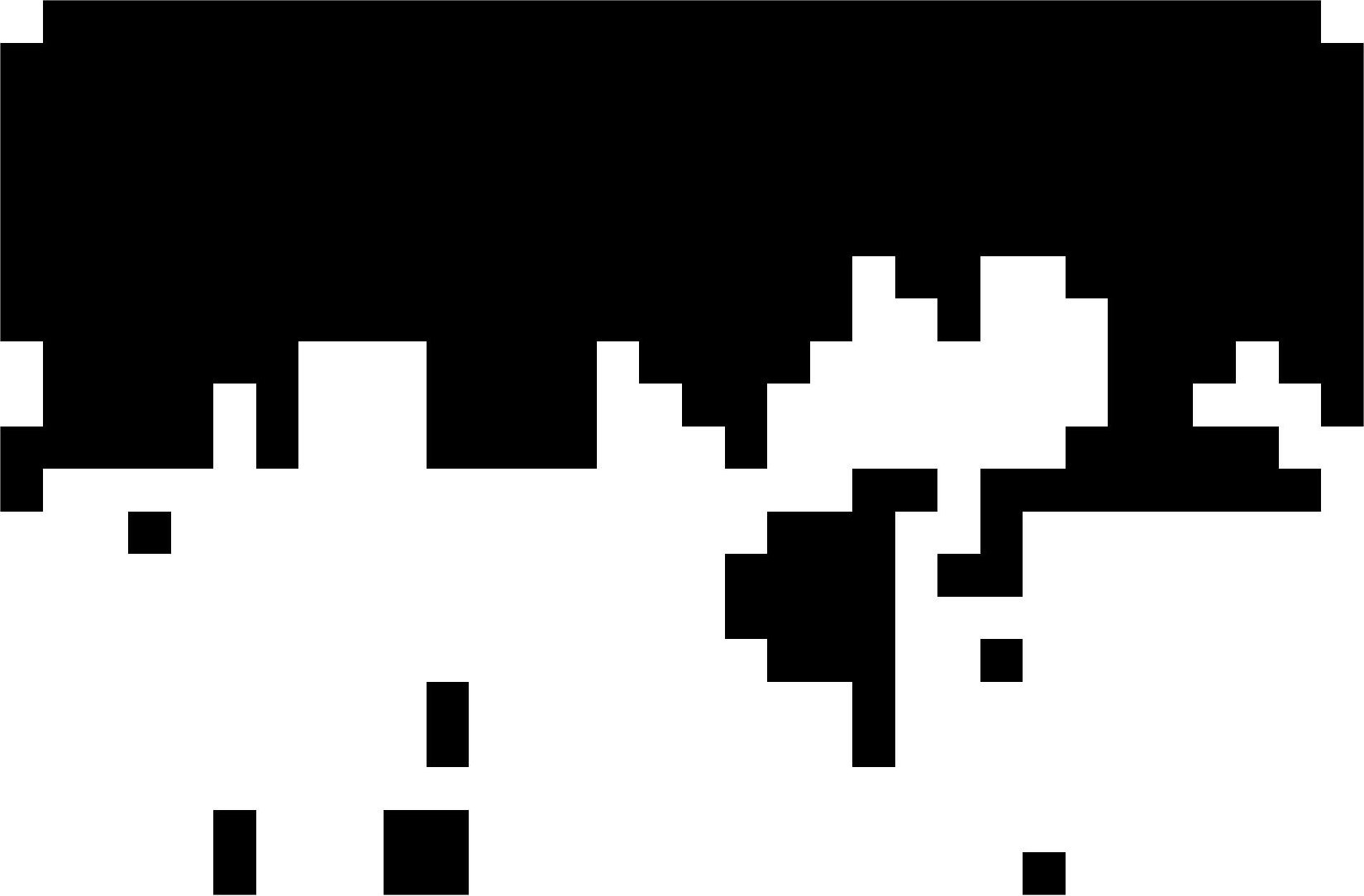}}
\end{subfigure}
\qquad
\begin{subfigure}[b]{\imagescale\textwidth}
\frame{\includegraphics[width=\textwidth]{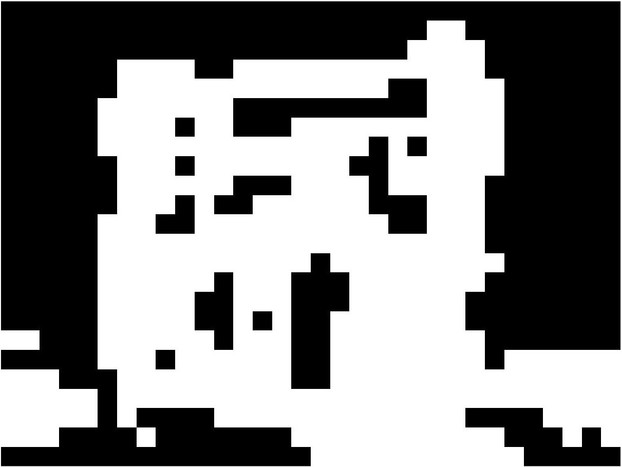}}
\end{subfigure}
\quad
\begin{subfigure}[b]{\imagescale\textwidth}
\frame{\includegraphics[width=\textwidth]{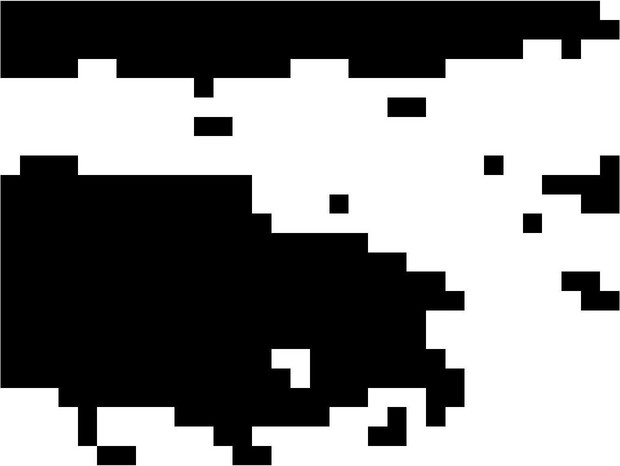}}
\end{subfigure}

\smallskip
(6) \begin{subfigure}[b]{\imagescale\textwidth}
{\includegraphics[width=\textwidth]{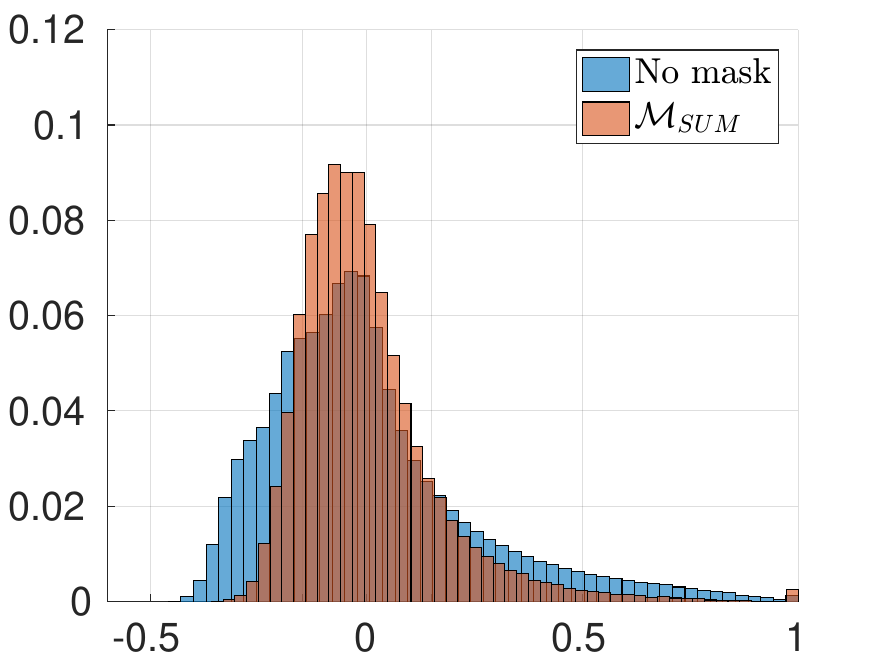}}
\end{subfigure}
\qquad
\begin{subfigure}[b]{\imagescale\textwidth}
{\includegraphics[width=\textwidth]{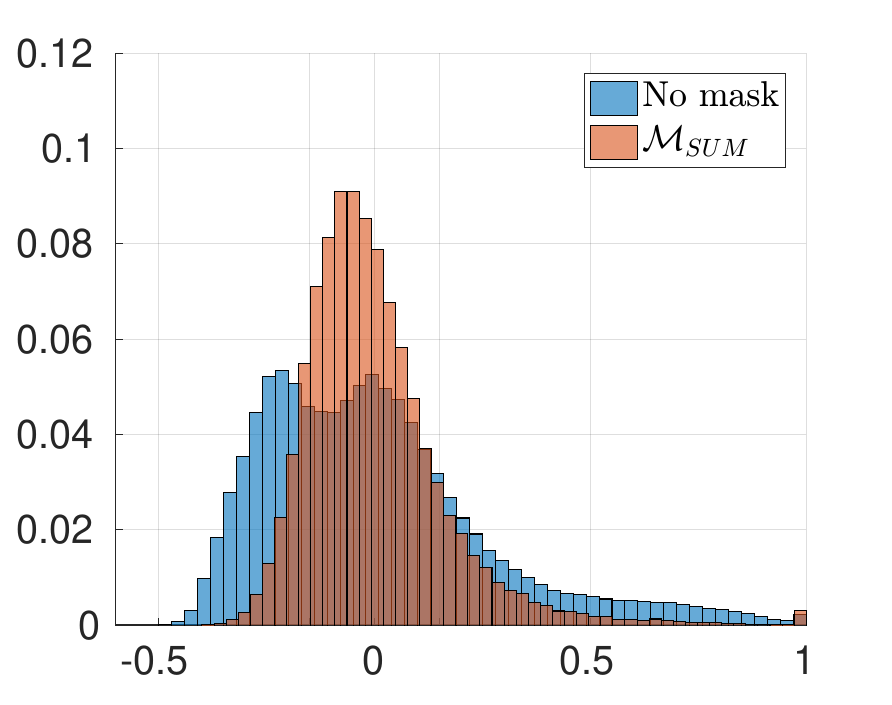}}
\end{subfigure}
\qquad
\begin{subfigure}[b]{\imagescale\textwidth}
{\includegraphics[width=\textwidth]{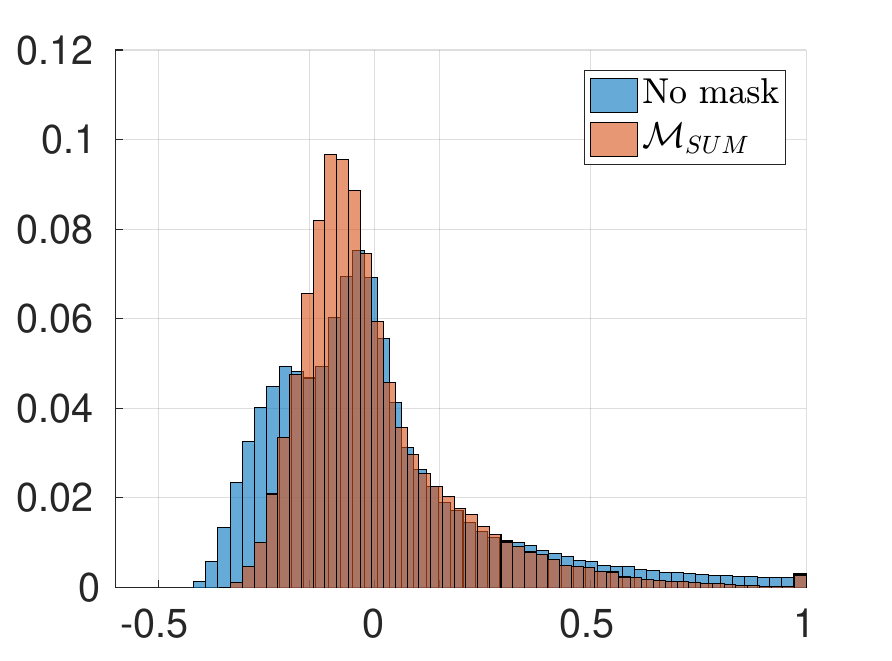}}
\end{subfigure}
\qquad
\begin{subfigure}[b]{\imagescale\textwidth}
{\includegraphics[width=\textwidth]{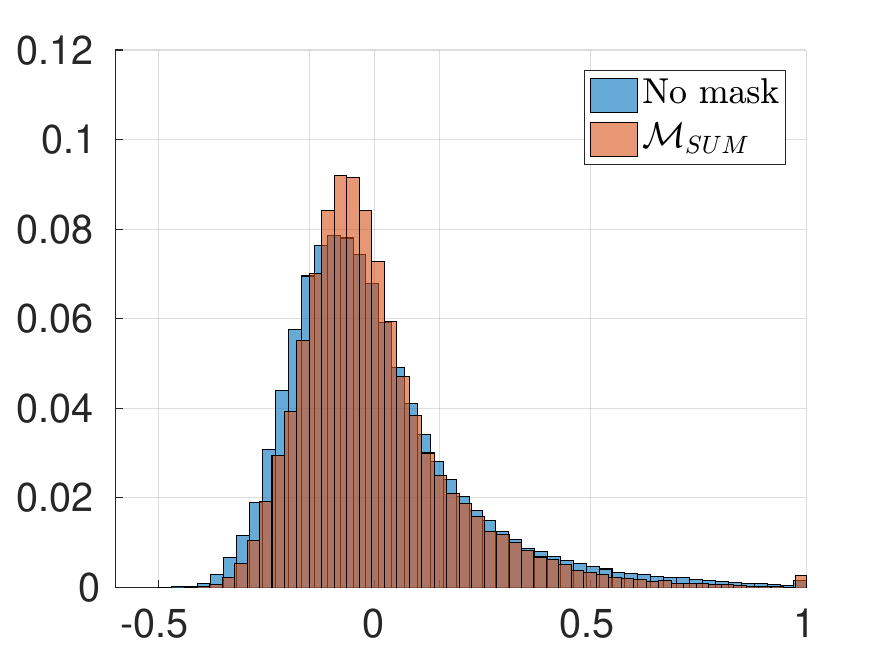}}
\end{subfigure}

\smallskip
(7) \begin{subfigure}[b]{\imagescale\textwidth}
\frame{\includegraphics[width=\textwidth]{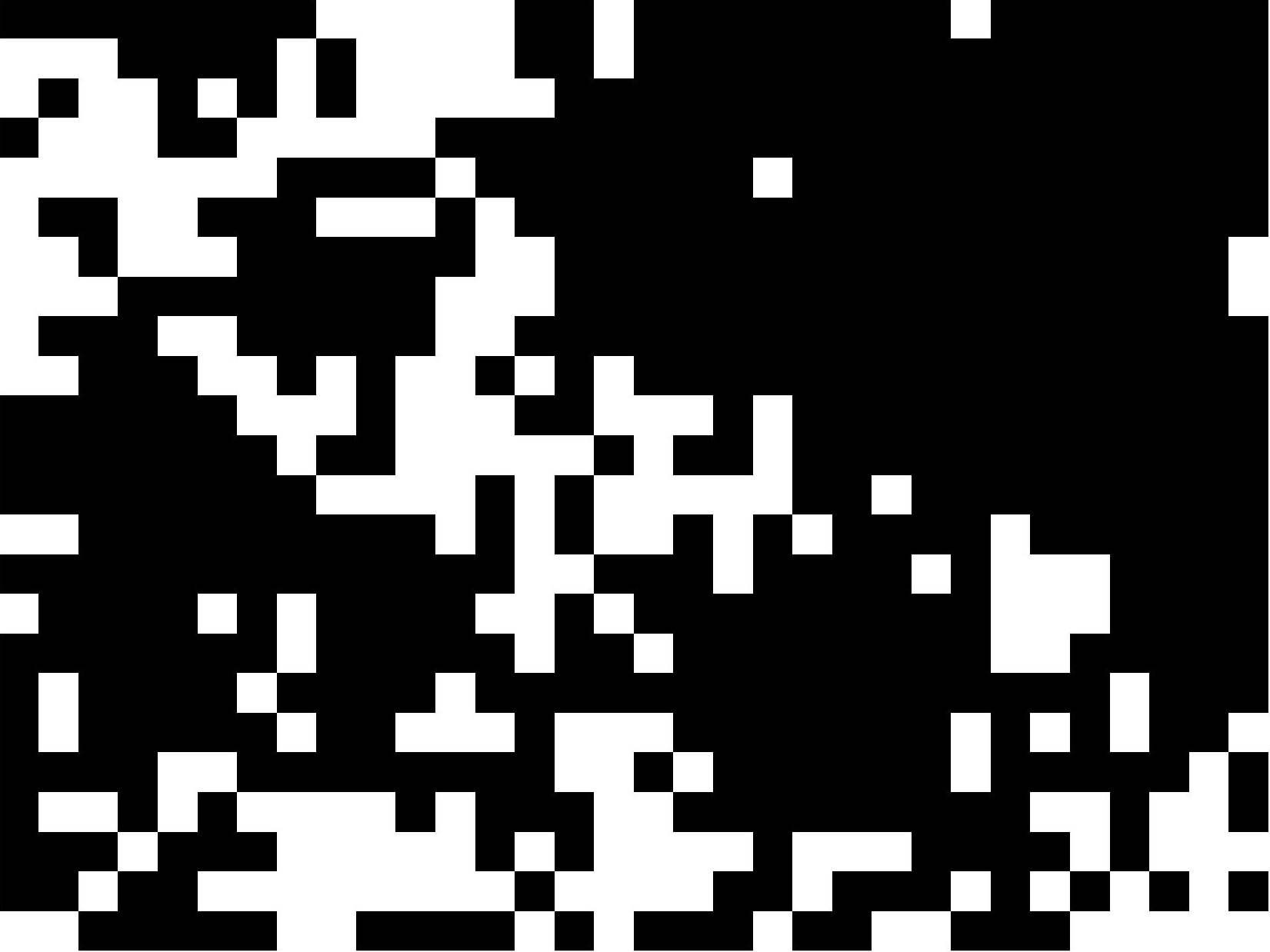}}
\end{subfigure}
\qquad
\begin{subfigure}[b]{\imagescale\textwidth}
\frame{\includegraphics[height=\imageheight,width=\textwidth]{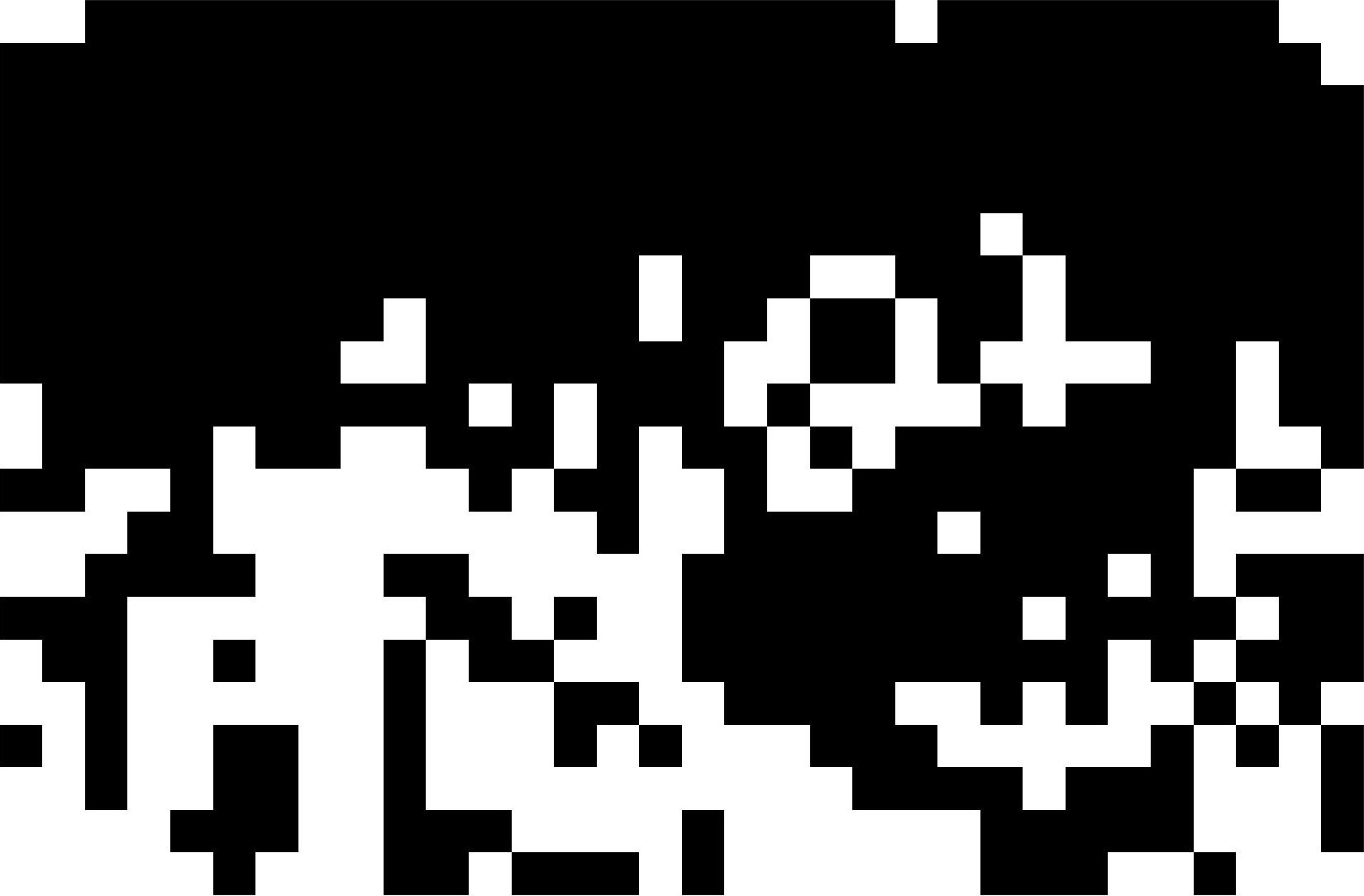}}
\end{subfigure}
\qquad
\begin{subfigure}[b]{\imagescale\textwidth}
\frame{\includegraphics[width=\textwidth]{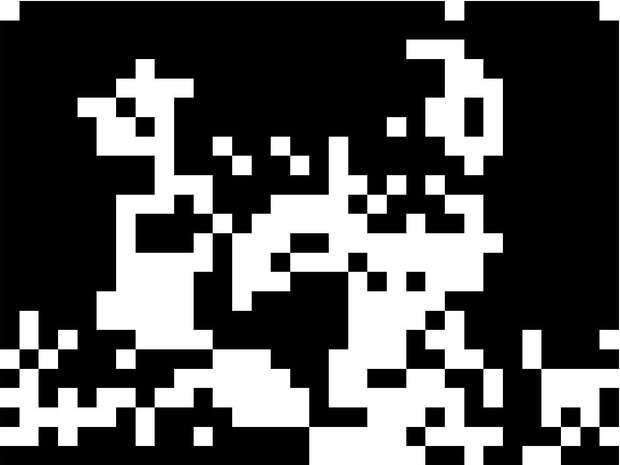}}
\end{subfigure}
\qquad
\begin{subfigure}[b]{\imagescale\textwidth}
\frame{\includegraphics[width=\textwidth]{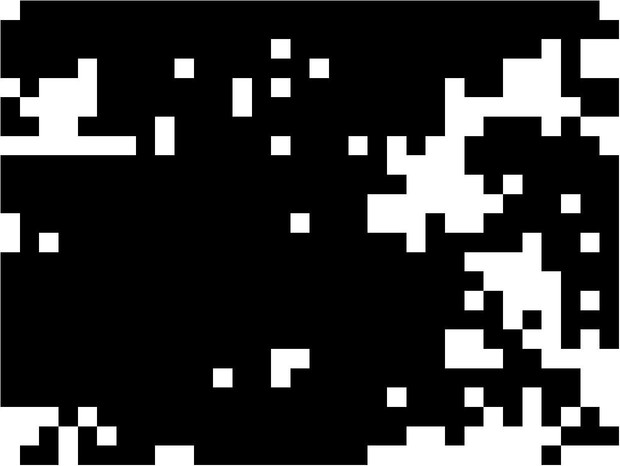}}
\end{subfigure}

\smallskip
(8) \begin{subfigure}[b]{\imagescale\textwidth}
{\includegraphics[width=\textwidth]{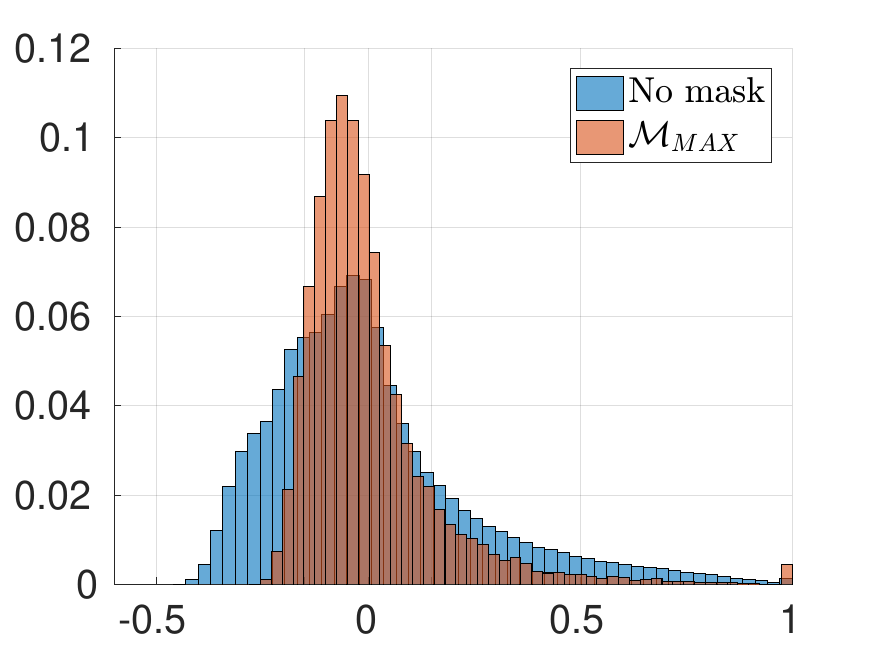}}
\end{subfigure}
\qquad
\begin{subfigure}[b]{\imagescale\textwidth}
{\includegraphics[width=\textwidth]{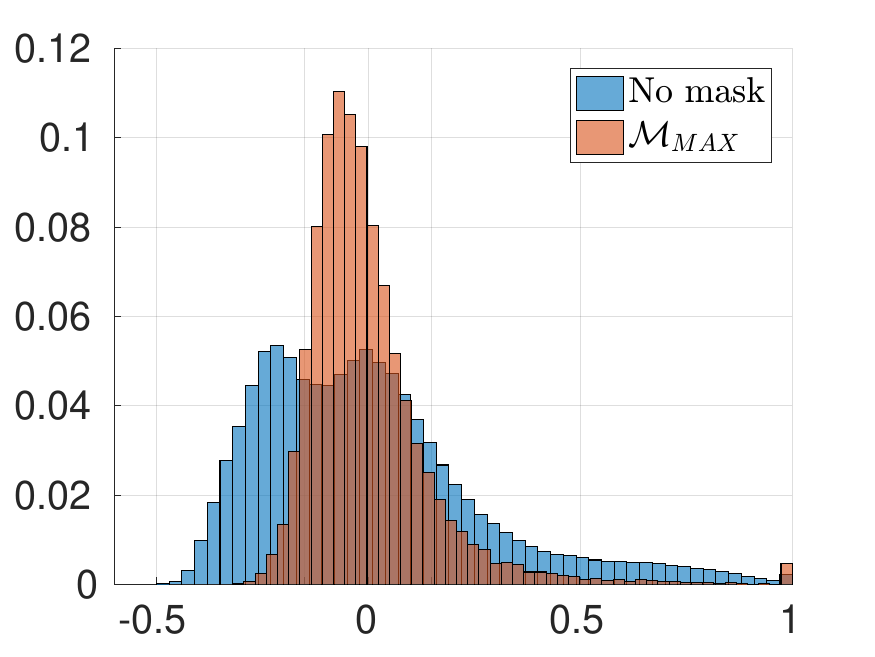}}
\end{subfigure}
\qquad
\begin{subfigure}[b]{\imagescale\textwidth}
{\includegraphics[width=\textwidth]{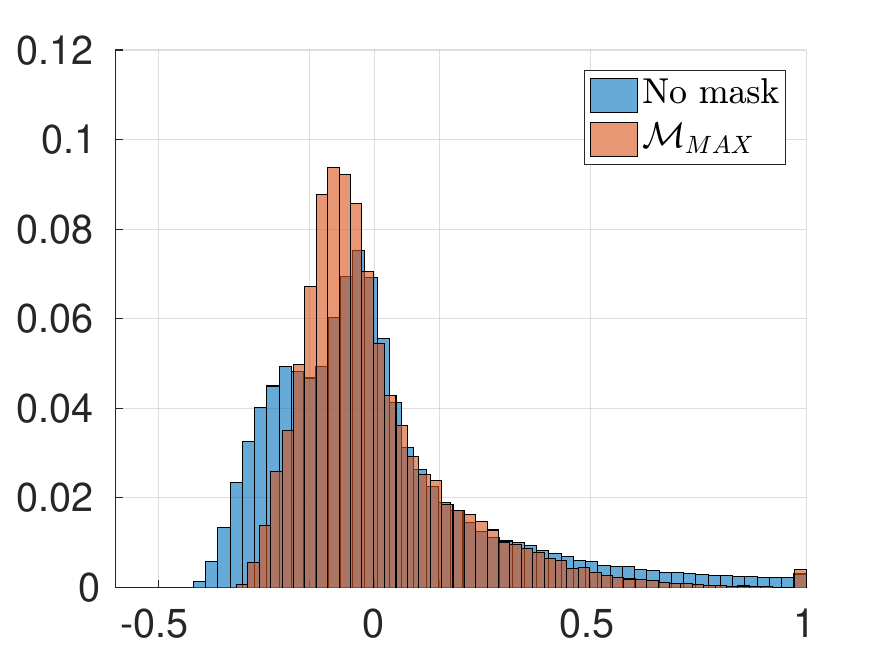}}
\end{subfigure}
\qquad
\begin{subfigure}[b]{\imagescale\textwidth}
{\includegraphics[width=\textwidth]{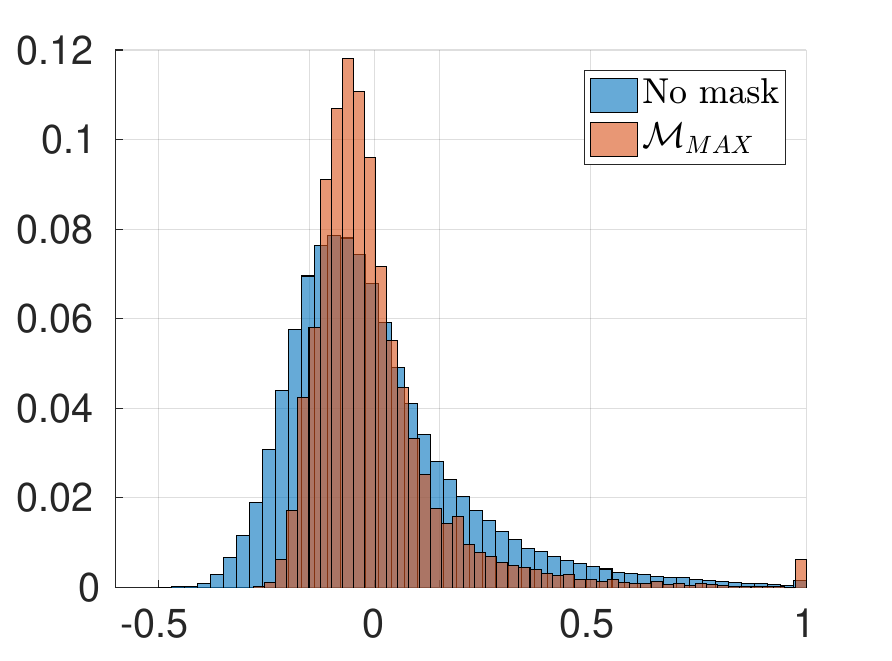}}
\end{subfigure}

\caption{Examples of SIFT/SUM/MAX-masks to select local conv. features. The first row shows the original images. The second row shows regions which are covered by SIFT features. The 3rd, 5th, and 7th rows respectively show the SIFT/SUM/MAX-masks of corresponding images (in the 1st row). The 4th, 6th, and 8th rows show the normalized histograms of covariances of sets of local conv. features with/without applying the SIFT/SUM/MAX-masks, respectively. }
\label{fig:mask-example}
\end{figure*}

In this section, firstly, we define the set of local deep conv. features which we work on throughout the paper (Section \ref{ssec:local_features}). We then propose in details the masking schemes to select a subset of discriminative local conv. features, including \textbf{\textit{SIFT-mask}}, \textbf{\textit{SUM-mask}}, and \textbf{\textit{MAX-mask}} (Section \ref{ssec:selective_fea}).
Finally, we provide in-depth analyses and experiments to qualitatively and quantitatively confirm the effectiveness of the proposed methods (Section \ref{ssec:effectiveness}). 

\subsection{Local deep convolutional features}
\label{ssec:local_features}
We consider a pre-trained CNN in which all fully connected layers are discarded. Given an input image $I$ of size $W_I \times H_I$ that is fed through a CNN, the 3D activation tensor of a conv. layer has the size of $W \times H \times K$ dimensions, where $K$ is the number of feature maps and $W \times H$ is the spatial resolution of a feature map. 
We consider this 3D tensor as a set $\mathcal{X}$ of $(W \times H)$ local features; each of them has $K$ dimensions. 
We denote $\mathcal{F}^{(k)}$ as $k$-th feature map with size of $W \times H$. 

\subsection{Selective features}
\label{ssec:selective_fea}
Inspired by the concept of finding the interest keypoints in the input images in traditional designs of hand-crafted features, we propose to select discriminative local deep conv. features.

We now formally propose different methods to compute a selection mask, i.e., a set of unique coordinates $\{(x,y)\}$ $(1 \le x \le W;1 \le y \le H)$ in the feature maps where local conv. features are retained.


\subsubsection{SIFT-Mask} 
\label{sssec:sift_mask}
In the image retrieval task, prior to the era of CNN, most previous works \cite{Temb, burstiness,vlad,FisherVector,neg_evidence,to_aggregate,revisitvlad} rely on SIFT \cite{SIFT_Lowe} features and its variant RootSIFT \cite{rootsift}. 
Although the gap between the SIFT-based representation and the semantic meaning of an image is still large, these early works have clearly demonstrated the capability of SIFT feature, especially in the potential of key-point detection. 
Fig. \ref{fig:mask-example} - Row (2) shows local image regions which are covered by SIFT. 
{We can obverse that SIFT features mainly cover the salient regions, i.e., buildings. This means that SIFT keypoint detector is capable of locating important regions of images. Hence, we propose to take the advantage of SIFT detector in combination with highly-discriminative local conv. features.} We will discuss more about the SIFT-mask in Section \ref{ssec:effectiveness}.

Specifically, let set $\mathcal{S}=\{(x^{(i)}, y^{(i)})\}_{i=1}^{n}$ be SIFT feature locations extracted from an $W_I \times H_I$ image; $1 \le x^{(i)} \le W_I$, $1 \le y^{(i)} \le H_I$. 
Based on the fact that conv. layers still preserve the spatial information of the input image \cite{R-MAC}, we select locations on the spatial grid $W \times H$ (of the feature map) which correspond to locations of SIFT key-points, i.e., 
\begin{equation}
\displaystyle{\mathcal{M}_{\text{SIFT}} = \left\{
\left(x_\text{SIFT}^{(i)}, y_\text{SIFT}^{(i)}\right) \right\}
\qquad i=1, \cdots, n;}
\end{equation}
where ${x_\text{SIFT}^{(i)}=\text{round}\left( \frac{x^{(i)}W}{W_I}\right)}$ and ${y_\text{SIFT}^{(i)}=\text{round}\left( \frac{y^{(i)}H}{H_I}  \right)}$, in which $\text{round}(\cdot)$ represents rounding to nearest integer. 
By keeping only locations $\mathcal{M}_{\text{SIFT}}$, we expect to remove ``background'' conv. features, while retaining ``foreground'' ones.


\subsubsection{MAX-Mask}
\label{sssec:max_mask} 
{It is widely known that each feature map contains the activations of a specific visual structure \cite{visual_CNN,RCNN}.} Hence, we propose to select the local conv. features which contain  high activations for all visual contents. In other words, we select the local features that capture the most prominent structures in the input images. These features are highly desirable to differentiate scenes.
In specific, we assess each feature map and select the location corresponding to the max activation value on that feature map. 
We formally define the selected locations $\mathcal{M}_{\text{MAX}}$ as follows:
\begin{equation}
\begin{split}
&\mathcal{M}_{\text{MAX}} = \left\{\left(x_\text{MAX}^{(k)}, y_\text{MAX}^{(k)}\right)\right\}\qquad k=1, \cdots, K;  \\
&\left(x_\text{MAX}^{(k)}, y_\text{MAX}^{(k)}\right) = \arg\max\limits_{(x,y)}\mathcal{F}_{(x,y)}^{(k)}.
\end{split}
\end{equation}

\subsubsection{SUM-Mask} {Departing from the MAX-mask idea, we propose a different masking method based on the motivation that a local conv. feature is more \textit{informative} if it gets excited in more feature maps, i.e., the sum on description values of a local feature is larger. }
By selecting local features having large values of sum, we can expect that those local conv. features are very informative about various local image structures \cite{visual_CNN}. The selected locations $\mathcal{M}_{\text{SUM}}$ is defined as follows:
\begin{equation}
\begin{split}
&\mathcal{M}_{\text{SUM}} = \left\{(x,y) \;|\; \Sigma_{(x,y)}^\mathcal{F} \ge \alpha \right\},\\
&\Sigma_{(x,y)}^\mathcal{F} = \sum_{k=1}^K\mathcal{F}_{(x,y)}^{(k)}, \qquad  \qquad \alpha=\text{median}(\Sigma^\mathcal{F}).
\end{split}
\vspace{-2em}
\end{equation}

\vspace{0.1cm}
\subsection{Effectiveness of the proposed masking schemes}
\label{ssec:effectiveness}
We now deeply analyze the effectiveness of the proposed masking schemes, in both qualitative and quantitative results. 

{
{SIFT detector \cite{SIFT_Lowe} is designed to detect interesting points which are robust with variations in scale, noise and illumination; therefore, these interesting points usually locate on high-contrast regions of images, e.g., corners. These regions also usually contain detail structures of the scenes which are necessary in differentiating scenes. While smooth regions, e.g., sky, road surfaces, are ignored as these regions are mainly background and contribute very little information. Hence, by using the SIFT-mask, we expect to select local conv. features at higher-contrast, i.e. potentially informative regions. However, there are two main issues when using the SIFT-mask: \textit{(i)} in cases of blurry images, the SIFT detector unsurprisingly fails to locate informative regions. 
\textit{(ii)} However, having too many interesting points also causes unexpected outcomes, which is known as the burstiness effect \cite{burstiness}, i.e., too many redundant local features are selected. For example, in Fig. \ref{fig:mask-example}-(2d)\footnote{Row (2) and column (d) of Fig. \ref{fig:mask-example}.} and \ref{fig:mask-example}-(3d), SIFT-mask includes almost all local features of the sea regions, which are obviously redundant}

On the other hand, SUM/MAX-masks perform much better when selecting just a few features at the sea region, i.e. Fig. \ref{fig:mask-example}-(5d) and \ref{fig:mask-example}-(7d) respectively, which are necessary to distinguish scenes with and without sea, and not to cause a serious burstiness effect which potentially makes the distinguishing different scenes with sea regions difficult. 
{In fact, the burstiness effect is the main reason explaining why SIFT-mask underperforms SUM/MAX-mask rather than due to SIFT-mask fails to select important regions, which is also confirmed by the two facts: \textit{(i)} SUM/MAX-masks are mainly subsets of SIFT-mask, and \textit{(ii)} applying SIFT-mask helps to improve performances (compared to no mask) which means that important regions have been selected, otherwise performances will drop.}
Note that the empirical results will be presented in Section \ref{sssec:framework}.
It is worth noting that, the burstiness effect on local conv. features is expectedly less severe since local conv. features have much larger receptive fields than those of SIFT features. Specifically, a local conv. feature extracted from $\mathtt{pool5}$ layer of AlexNet \cite{Alexnet} and VGG16 \cite{VGG} have the receptive fields of $195\times 195$ and $212\times 212$ respectively. We will further investigate this effect in Section \ref{ssec:post_processing}.

{Comparing SUM-mask and MAX-mask, which are computed from learned features, they both have the capability of detecting important regions based on the responded activation of regions. However, their principles of selecting local features are different. In particular, given prominent regions, the corresponding local conv. features of those regions are usually highly activated. As a result, the sums on those features are larger. This fact explains why SUM-mask more densely selects local conv. features at prominent regions. However, as the receptive fields of neighbouring features are largely overlapping, they are likely to contain similar information, i.e., redundant. On the other hand, MAX-mask only selects the features which have highest activation values. Hence, we expect MAX-mask can select the best features for representing the visual structures of prominent regions. As a result, we minimize the chance of selecting multiple similar local features.}
}

\begin{wrapfigure}{r}{0.6\textwidth}
\vspace{-1em}
\centering
\begin{subfigure}[b]{0.28\textwidth}
\includegraphics[width=\textwidth]{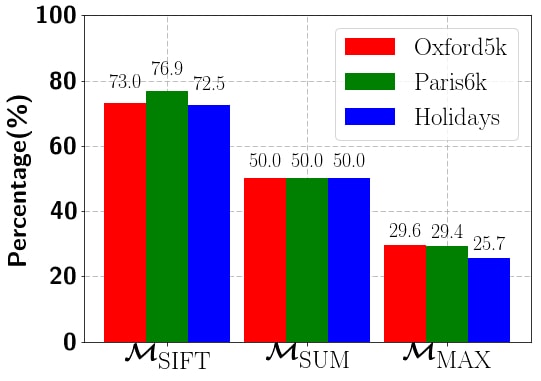}
\caption{}
\label{sfig:num_local_fea}
\end{subfigure}
\begin{subfigure}[b]{0.28\textwidth}
\includegraphics[width=\textwidth]{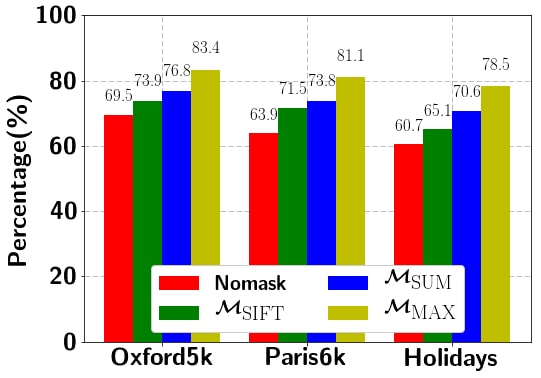}
\caption{}
\label{sfig:std}

\end{subfigure}

\caption{Fig. \ref{sfig:num_local_fea}: The averaged percentage of remaining  local conv. features after applying masks.
Fig. \ref{sfig:std}: The averaged percentage of the covariance values in the range of $[-0.15, 0.15]$.
}
\label{fig:hist}
\vspace{-1em}
\end{wrapfigure}




Besides, we quantitatively evaluate the effectiveness of our proposed masking schemes in eliminating redundant local conv. features.
Firstly, Fig. \ref{sfig:num_local_fea} shows the averaged percentage  of the remaining local conv. features after applying our proposed masks on \textit{Oxford5k} \cite{oxford5k}, \textit{Paris6k} \cite{paris6k}, and \textit{Holidays} \cite{holiday} datasets (Section \ref{ssec:datasets}). Note that local conv. features are extracted from $\mathtt{pool5}$ layer of the pre-trained VGG \cite{VGG} with the input image size of $\max(W_I,H_I)=1024$. Apparently, SIFT/SUM/MAX-masks remove large numbers of local conv. features, about 25\%, 50\%, and 70\% respectively.

{In addition, we present the normalized histograms of covariances of selected local conv. features after applying different masks in Fig. \ref{fig:mask-example}-Row 4th, 6th, and 8th. To compute the covariances, we first $l2$-normalize local conv. features, and then compute the dot products for all pairs of features. For easy comparison, the normalized histograms of covariances of all available local conv. features (i.e., before masking) are included. 
We can clearly observe that the distributions of covariances after applying masks have much higher peaks around 0 and have smaller tails than those without applying masks. 
This indicates that the masks are helpful in reducing correlation between features.
Additionally, Fig. \ref{sfig:std} shows the averaged percentage of $l2$-normalized feature pairs whose dot products are within the range of $[-0.15, 0.15]$. The chart shows that the selected features are more uncorrelated. }
In summary, Fig. \ref{fig:hist} shows that the proposed masking schemes are effective in removing a large proportion of redundant local conv. features. As a result, we can select a better representative subset of local conv. features. Furthermore, as the number of features is reduced, the computational cost is also considerably reduced, especially for the subsequent embedding and aggregating steps.

\section{Framework: Embedding and aggregating on selective convolution features}
\label{sec:frame}
In this section, we introduce the completed framework which takes a set of local deep conv. features to compute the final image representation. 

\subsection{Pre-processing} 
Given a set $\mathcal{X}_\mathcal{M}=\{\mathbf{x}_{(x,y)}~|~(x,y)\in \mathcal{M}_*\}$, where $\mathcal{M}_* \in \{\mathcal{M}_\text{SUM}, \mathcal{M}_\text{MAX}, $ $ \mathcal{M}_\text{SIFT}\}$, of selective $K$-dimensional local conv. features belonged to the set,
we apply the principal component analysis (PCA) to compress local conv. features to a lower dimension $d$: $\mathbf{x}^{(d)} =M_{\text{PCA}}\mathbf{x}$, where $M_{\text{PCA}}$ is the PCA-matrix. 
Applying PCA for dimension reduction can be very beneficial for two reasons.
Firstly, using low-dimensional local features can help to produce compact final image representations as done in recent state-of-the-art image retrieval methods \cite{cnn_max_pooling,R-MAC,finetune_hard_samples}. 
Secondly, applying PCA could be helpful in removing noise and redundancy; hence, enhancing the discrimination. 
The compressed features are subsequently $l2$-normalized.

\subsection{Embedding}
\label{ssec:embedding}

We additionally aim to boost the discrimination power of the selective local conv. features.  
This task can be accomplished by embedding the local features to a high-dimensional space: $\mathbf{x}\mapsto \phi(\mathbf{x})$, using state-of-the-art embedding methods: \textit{Fisher vector} -- FV~\cite{FisherVector}, \textit{vector of locally aggregated descriptors} -- VLAD~\cite{vlad}, \textit{triangulation embedding} -- Temb~\cite{Temb}, \textit{function appoximation-based embedding} -- F-FAemb~\cite{F-FAemb}. It is worth noting that while in \cite{cnn_max_pooling}, the authors mentioned that local conv. features are already discriminative; hence the embedding step is not necessary.
However, in this work, we find that embedding the \textit{selected} features to higher dimension significantly improves their discriminability.  

\subsection{Aggregating} 
\label{ssec:aggregating}
Let $\mathbf{V}_i = [\phi(\mathbf{x}_1^i),\cdots, \phi(\mathbf{x}_{n_i}^i)]$ be an $D\times n_i$ matrix that contains $n_i$ $D$-dimensional embedded local descriptors of $i$-th image. 
In earlier works, the two common methods to aggregate a set of local features to a single global one are max-pooling ($\psi_\text{m}$) and sum/average-pooling ($\psi_\text{s}/\psi_\text{a}$).
Recently, H. J{\'e}gou et al. \cite{Temb} introduced \textbf{\textit{democratic aggregation}} $(\psi_\text{d})$ method applied to image retrieval problem. The fundamental idea of democratic aggregation is to equalize the similarity between each local features and the aggregated representation. Note that, concurrently, Murray and Perronnin \cite{gmp} proposed \textbf{\textit{Generalized Max Pooling (GMP)}} ($\psi_\text{GMP}$), which shares the similar idea with \textit{democratic aggregation}. 
Democratic aggregation can be directly applied on
various embedded features, e.g., FV \cite{FisherVector}, VLAD \cite{vlad}, Temb \cite{Temb}, F-FAemb \cite{F-FAemb}. Moreover, when working with embedded SIFT features, this aggregation method has been shown to clearly outperform both max-pooling and sum/average-pooling \cite{Temb}. Noted that democratic aggregation requires local features to be $l2$-normalized.

\subsection{Post-processing}
\label{ssec:post_processing}

\textbf{\textit{Power-law normalization (PN).}} The \textit{burstiness} of visual elements \cite{burstiness} is the phenomenon that numerous descriptors are almost similar within an image. The {burstiness} can severely impact the similarity measure between two images. 
An effective solution to the burstiness issue is to apply PN \cite{power-norm} to and subsequently $l2$-normalize \cite{Temb} the aggregated features $\psi$.
The PN formulation is defined as $PN(x)=\text{sign}(x)|x^\alpha|$, where $0 \le \alpha \le 1$ \cite{power-norm}.

\begin{wrapfigure}{tr}{0.6\textwidth}
\vspace{-1.5em}
\centering
\begin{subfigure}[b]{0.28\textwidth}
\includegraphics[width=\textwidth]{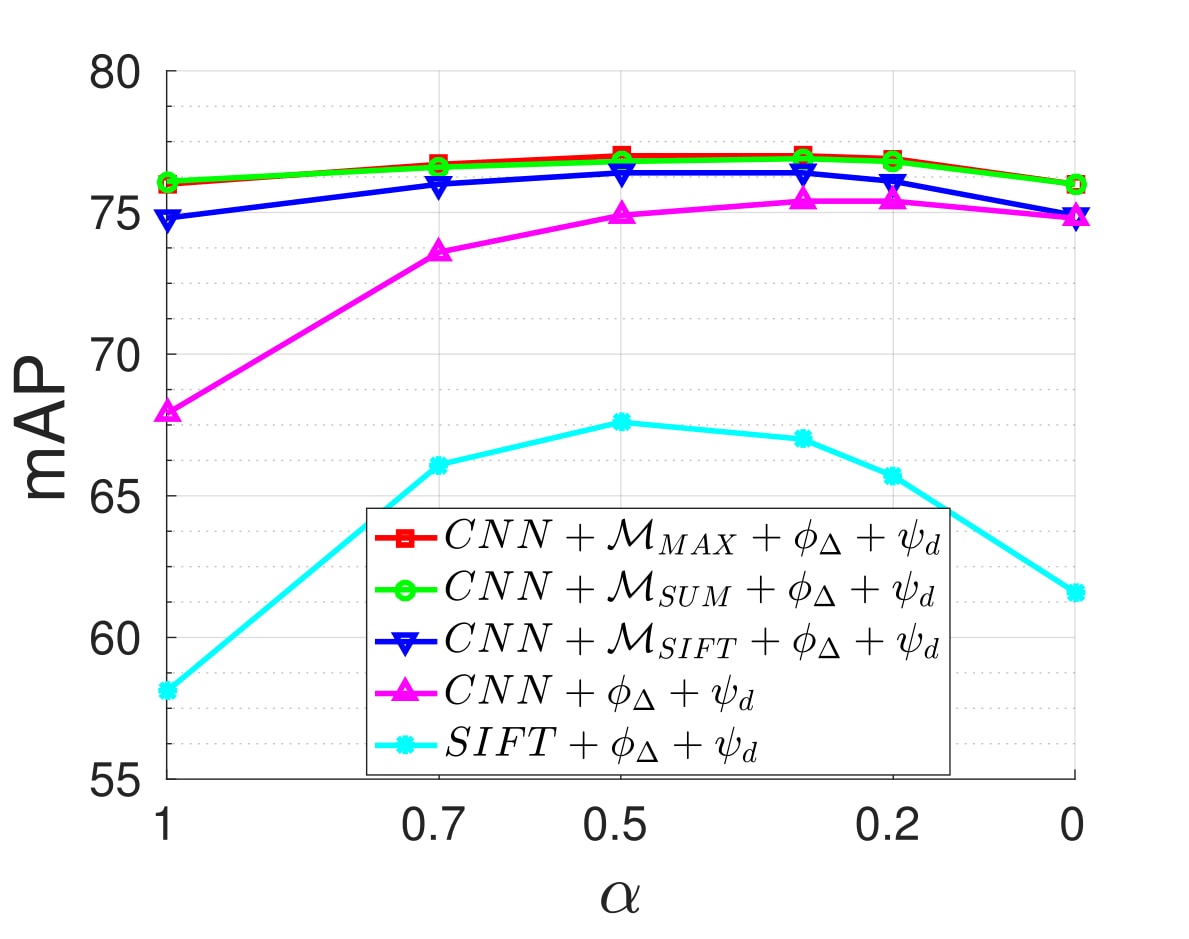}
\caption{\textbf{\textit{Oxford5k}}}
\end{subfigure}
\begin{subfigure}[b]{0.28\textwidth}
\includegraphics[width=\textwidth]{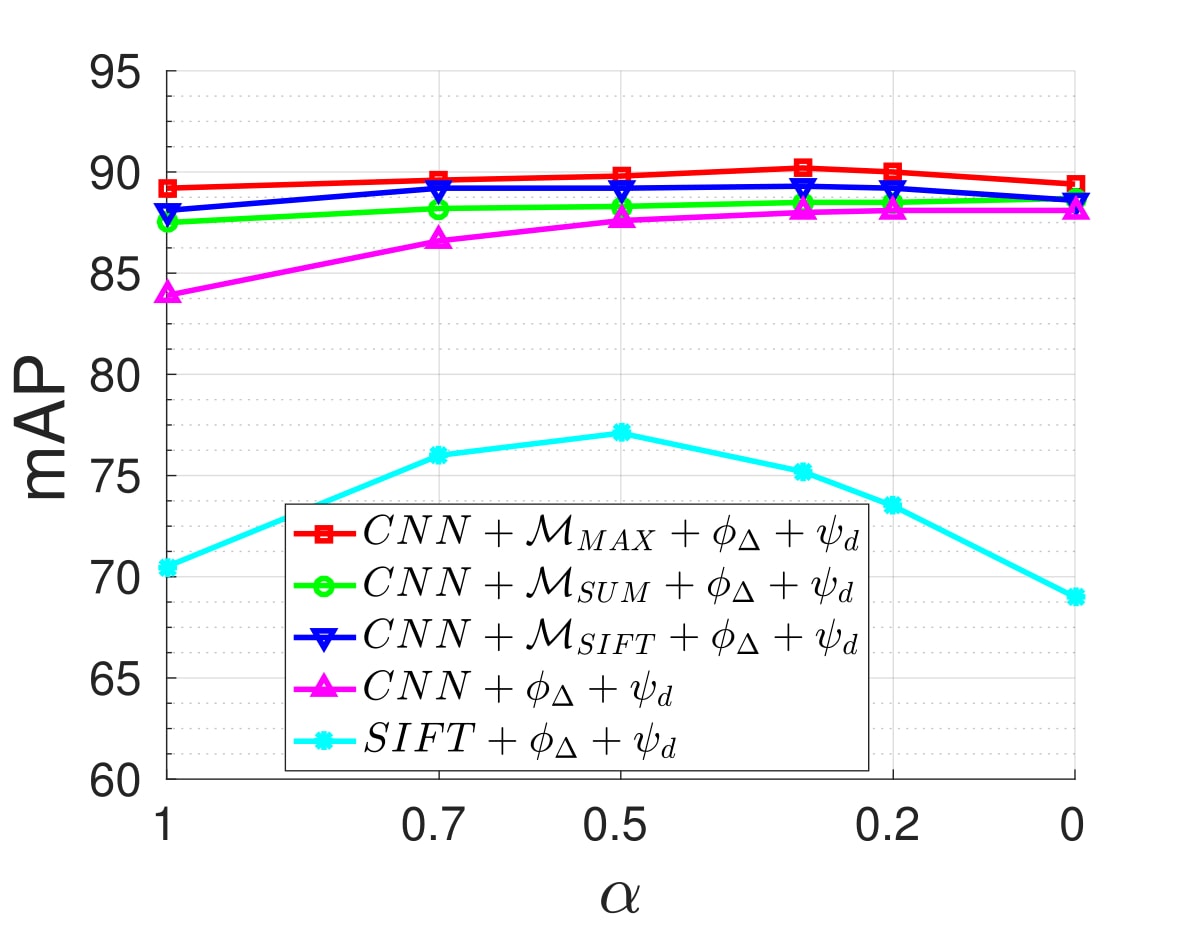}
\caption{\textbf{\textit{Holidays}}}
\end{subfigure}
\caption{Impact of power-law normalization factor $\alpha$ on retrieval performance.
Following the setting in \cite{Temb}, we set $d=128$ and $|\mathcal{C}|=64$ for both SIFT and conv. features. The local conv. features are extracted from $\mathtt{pool5}$ layer of the pre-trained VGG \cite{VGG}.}
\label{fig:norm-effect}
\vspace{-1em}
\end{wrapfigure}

By the best of our knowledge, no previous work has re-studied the \textit{burstiness} phenomena on the local conv. features. 
Fig. \ref{fig:norm-effect} shows the effect of PN on local conv. features using various proposed masking schemes. The figure shows that the burstiness still happens on local conv. features ($CNN+\phi_\Delta+\psi_\text{d} $), as the retrieval performance changes as $\alpha$ varies.
However, we additionally observe that the burstiness on conv. features is much weaker than on SIFT features ($SIFT+\phi_\Delta+\psi_\text{d}$). 
More importantly, the proposed SIFT/SUM/MAX-masks clearly mitigate the burstiness phenomena: the performances achieved by $CNN+\mathcal{M}_\text{MAX/SUM/SIFT}+ \phi_\Delta+\psi_\text{d}$ are stable as $\alpha$ varies. This confirms the effectiveness of the proposed masking schemes in removing redundant local features. Following previous works, $\alpha$ is set at 0.5 for all later experiments, unless stated otherwise.

\textbf{\textit{Rotation normalization and dimension reduction (RN).}} 
Besides the visual burstiness, frequent co-occurrences issue is also an important limitation. Fortunately, this effect can be easily addressed by whitening the data.


\subsection{Hashing function}
\label{sssec:hash}
In the large scale image retrieval problem, binary hashing, where images are represented by a $L$-bit binary codes, is an attractive approach because the binary representations allow the fast searching and sufficient storage.

There is a wide range of hashing methods have been proposed in the literature, in both unsupervised and supervised~\cite{Grauman_review,survey_learn2hash}. Although supervised hashing methods usually outperform unsupervised hashing methods on some specific retrievals in which the data is labeled, they are not suitable for the general image retrieval (which is focused in this work). That is because in the general image retrieval, the label of an image is not well defined. Most of general image retrieval benchmarks, e.g., Holidays, Oxford5k, Paris6k, does not have labeled training data. On the other hand, the unsupervised hashing is well suitable for the general image retrieval task. Unsupervised hashing methods do not require the data label for training. Most of unsupervised hashing methods tries to preserve the geometric structure of data by using reconstruction criterion~\cite{ITQ,BA_CVPR15,SAH,do2016learning} or directly preserving the distance similarity between  samples~\cite{SpH}. 
By above reasons, we propose to cascade a state-of-the-art unsupervised hash function, i.e., Iterative Quantization (ITQ) \cite{ITQ}, K-mean Hashing (KMH) \cite{kmeanHash}, or Relaxed Binary Autoencoder (RBA) \cite{SAH},  into the framework to further binarize the real-valued aggregated representations to binary representations.

\smallskip
The overview of our proposed framework is shown in Fig. \ref{fig:framework}. In the next section, we will conduct extensive experiments to evaluate the framework in both cases: real-valued global representations (i.e., without the hash function in the framework), and binary global representations (i.e., with the hash function in the framework).
\section{Evaluation}
\label{sec:exp}
In this section, we conduct a wide range of experiments to comprehensively evaluate the  proposed framework on six standard image retrieval benchmark datasets, including 
Oxford5k dataset \cite{oxford5k}, 
Paris6k dataset \cite{paris6k}, 
INRIA Holidays \cite{holiday} dataset, 
Oxford105k dataset \cite{oxford5k}, 
Paris106k dataset \cite{paris6k}, 
and Holidays+Flickr1M dataset \cite{herve_eccv2008}.

\subsection{Datasets, evaluation protocols, and implementation notes}
\label{ssec:datasets}

\textbf{Oxford Buildings dataset: } The \textit{Oxford5k} dataset \cite{oxford5k} consists of 5,063 images of buildings and 55 query images corresponding to 11 distinct buildings in Oxford. Each query image contains a bounding box indicating the region
of interest. Following the standard practice \cite{Temb,F-FAemb,R-MAC,deep_img_retr}, we use the cropped query images based on provided bounding boxes.

\textbf{Paris dataset: } The \textit{Paris6k} dataset \cite{paris6k} consists of 6412 images of famous landmarks in Paris. Similar to \textit{Oxford5k}, this dataset has 55 queries corresponding to 11 landmarks. 
We also use provided bounding boxes to crop the query images accordingly. 

\textbf{INRIA Holidays dataset:} The \textit{Holidays} dataset \cite{holiday} contains 1,491 images corresponding to 500 scenes. The query image set consists of one image from each scene. Following \cite{neuralcode,cnn_max_pooling,CroW}, we manually rotate images (by $\pm 90$ degrees) to fix the incorrect image orientation. 

\textbf{Oxford105k and Paris106k datasets: } We additionally combine \textit{Oxford5k} and \textit{Paris6k} with 100k Flickr images~\cite{oxford5k} to form larger databases, named \textit{Oxford105k} and \textit{Paris106k} respectively. 
The new databases are used to evaluate retrieval performance at a larger scale.

\textbf{Holidays+Flickr1M: } In order to evaluate the retrieval on a very large scale, we merge Holidays dataset with 1M negative images downloaded from Flickr \cite{herve_eccv2008}, forming the Holidays+Flickr1M dataset. This dataset allows us to evaluate real-like scenarios of the proposed framework.

\textbf{Evaluation protocols:} Follow the state of the art \cite{Temb,F-FAemb,netvlad,cnn_max_pooling,finetune_hard_samples,deep_img_retr}, the retrieval performance is measured by mean average precision (\textbf{\textit{mAP}}) over the query sets. Additionally, the \textit{junk} images
are removed from the ranking.

\begin{wraptable}{r}{0.6\textwidth}
\vspace{-1.2em}
\footnotesize
\caption{Notations and their corresponding meanings. 
}
\label{tb:notation_sum}
\centering
\begin{tabular}{|c|l||c|l|}
\hline
Notations & Meanings & Notations & Meanings \\
\hline
$\mathcal{M}_\text{SIFT}$ & SIFT-mask & $\psi_\text{a}$ & Average-pooling\\
$\mathcal{M}_\text{SUM}$ & SUM-mask & $\psi_\text{s}$ & Sum-pooling\\
$\mathcal{M}_\text{MAX}$ & MAX-mask & $\psi_\text{d}$ & Democratic-pooling \cite{Temb}\\
$\phi_\text{FV}$ & FV \cite{FisherVector} & $\phi_\text{VLAD}$ & VLAD \cite{vlad} \\
$\phi_\Delta$ & Temb \cite{Temb} & $\phi_\text{F-FAemb}$ & F-FAemb \cite{F-FAemb} \\
$\mathcal{C}$ & Codebook\footnotemark & $d$ & Retained PCA dim. \\
$D$ & Final dim. & &\\
\hline
\end{tabular}
\vspace{-0.5em}
\end{wraptable}

\textbf{Implementation notes:} In the image retrieval task, to avoid overfitting, it is important to use held-out datasets (training set) to learn all necessary parameters \cite{cnn_max_pooling, finetune_hard_samples,deep_img_retr}. Following standard settings in the literature~\cite{Temb,F-FAemb,R-MAC,cnn_max_pooling}, we use the set of 5,000 Flickr images \cite{oxford5k}
\footnote{We randomly select 5,000 images from the 100k Flickr image set \cite{oxford5k}.} 
as the training set to learn parameters for \textit{Holidays} and \textit{Holidays+Flick1M}. The \textit{Oxford5k} is used as the learning set for \textit{Paris6k} and \textit{Paris106k}, while the \textit{Paris6k} is used as the learning for \textit{Oxford5k} and \textit{Oxford105k}
%
%
%
%
For fair comparison, following recent works \cite{R-MAC, cnn_max_pooling, finetune_hard_samples, deep_img_retr}, we use the pretrained VGG16 \cite{VGG} (with Matconvnet toolbox \cite{matconvnet}) to extract deep conv. features.
In addition, all images are resized so that the maximum dimension is 1,024 while preserving aspect ratios before fed into the CNN. 
We utilize the VLFeat toolbox \cite{vlfeat} for SIFT detector. 
For clarity, the notations are summarized in Table \ref{tb:notation_sum}. The implementation of the proposed framework is available at \url{https://github.com/hnanhtuan/selectiveConvFeature}.

\footnotetext{For FV method, the codebook is learned by Gaussian Mixture Model. For VLAD, Temb, and F-FAemb methods, the codebooks learned by K-means.}

\subsection{Effects of parameters}
\subsubsection{Frameworks}
\label{sssec:framework}
In this section, we conduct experiments to comprehensively evaluate various embedding and aggregating methods in combination with different proposed masking schemes. 
Note that, we follow \cite{F-FAemb} to decompose the embedding and aggregating steps of VLAD and FV methods. This allows us to utilize the state-of-the-art aggregations (e.g., democratic pooling \cite{Temb}).

\begin{wraptable}{r}{0.6\textwidth}
\vspace{-0.8em}
\small
\caption{Configurations of different embedding methods.}
\label{tb:emb-setting}
\centering
\begin{tabular}{|l|c|c|c|}
\hline
\textbf{Methods} & $d$ & $|\mathcal{C}|$ & $D$ \\
\hline
FV \cite{FisherVector} & 48 & 44 & $2\times d\times |\mathcal{C}|=4224$\\
VLAD \cite{vlad} & 64 & 66 & $d\times |\mathcal{C}|=4224$  \\
T-emb \cite{Temb} & 64 & 68 & $d\times |\mathcal{C}|-128=4224$ \\
F-FAemb \cite{F-FAemb}\footnotemark & 32 & 10 & $\displaystyle{\frac{(|\mathcal{C}|-2)\times d\times (d+1)}{2}=4224}$\\
\hline
\end{tabular}
\vspace{-0.5em}
\end{wraptable}
\footnotetext{Instead of removing the first $d(d+1)/2$ components as in original design \cite{F-FAemb}, we remove the first $d(d+1)$ components of the features after aggregating step (Section \ref{ssec:aggregating}) as this generally achieves better performances.}

 In order to have a fair comparison among different combinations, we empirically set the visual codebook size-$|\mathcal{C}|$ and the number of retained PCA components-$d$ (of local conv. features) such that the produced final aggregation vectors of different methods have the same dimensionality-$D$. These parameters are presented in Table \ref{tb:emb-setting}. 

We report the comparative results on \textit{Oxford5k}, \textit{Paris6k}, and \textit{Holidays} datasets in Table \ref{tb:compare_pooling_framework}. 
The main observations from Table \ref{tb:compare_pooling_framework} are: \textit{(i)} democratic pooling is clearly better than sum/max-pooling, \textit{(ii)} our proposed masking schemes consistently boost performance for all embedding and aggregating frameworks, and finally \textit{(iii)} the MAX-mask outperforms the SUM/SIFT-masks, while the performance gains of SUM-mask and SIFT-mask are comparable.
At the comparison dimensionality of $4224-D$, the two frameworks $\phi_\Delta+\psi_\text{d}$ and $\phi_\text{F-FAemb}+\psi_\text{d}$ achieve comparable performances for various masking schemes and datasets. Hence, we choose $\mathcal{M}_* + \phi_\Delta+\psi_\text{d}$ as our default framework for analyzing other parameters.

\begin{table}[!t]
\small
\caption{Comparison of different frameworks. For simplicity, we do not include the notations for post-processing steps (PN and RN). The \textbf{``Bold''} values indicate the best performances in each masking scheme and the \underline{``Underline''} values indicate the best performances across all settings.}
\label{tb:compare_pooling_framework}
\centering
\begin{tabular}{|c|l|c|c|c|c|}
\hline
& \textbf{Frameworks} & $\mathcal{M}_\text{MAX}$ & $\mathcal{M}_\text{SUM}$ & $\mathcal{M}_\text{SIFT}$ & None \\
\hline
\multirow{6}{*}{\rotatebox[origin=c]{90}{\textbf{Oxford5k}}} & $\phi_\text{FV}+\psi_\text{a}$ & 67.8 & 65.1 &  65.5 & 59.5 \\
& $\phi_\text{FV}+\psi_\text{d}$ & 72.2 & 71.8 & 72.0 & 69.6 \\
& $\phi_\text{VLAD}+\psi_\text{s}$ & 66.3 & 65.6 & 66.4 & 65.1 \\
& $\phi_\text{VLAD}+\psi_\text{d}$ & 69.2 & 70.5 & 71.3 & 69.4 \\
& $\phi_\Delta+\psi_\text{d}$ & \underline{\textbf{75.8}} & \textbf{75.7} & \textbf{75.3} & 73.4 \\
& $\phi_\text{F-FAemb}+\psi_\text{d}$ & 75.2 & 74.7 & 74.4 & \textbf{73.8} \\
\hline
\multirow{6}{*}{\rotatebox[origin=c]{90}{\textbf{Paris6k }}} & $\phi_\text{FV}+\psi_\text{a}$ & 78.4 & 76.4 & 75.8 & 68.0 \\
& $\phi_\text{FV}+\psi_\text{d}$ & 84.5 & 82.2 & 82.4 & 76.9 \\
& $\phi_\text{VLAD}+\psi_\text{s}$ & 77.7 & 74.5 & 76.0 & 73.2 \\
& $\phi_\text{VLAD}+\psi_\text{d}$ & 80.3 & 79.5 & 81.3 & 79.3 \\
& $\phi_\Delta+\psi_\text{d}$ & \underline{\textbf{86.9}} & 84.8 & 85.3 & \textbf{83.9} \\
& $\phi_\text{F-FAemb}+\psi_\text{d}$ & 86.6 & \textbf{85.9} & \textbf{85.6} & 82.9 \\
\hline
\multirow{6}{*}{\rotatebox[origin=c]{90}{\textbf{Holidays }}} & $\phi_\text{FV}+\psi_\text{a}$ & 83.2 & 80.0 & 81.5 & 78.2 \\
& $\phi_\text{FV}+\psi_\text{d}$ & 87.8 & 86.7 & 87.1 & 85.2 \\
& $\phi_\text{VLAD}+\psi_\text{s}$ & 83.3 & 82.0 &  83.6 & 82.7 \\
& $\phi_\text{VLAD}+\psi_\text{d}$ & 85.5 & 86.4 & 87.5 & 86.1 \\
& $\phi_\Delta+\psi_\text{d}$ & \underline{\textbf{89.1}} & 88.1 & \textbf{88.6} & \textbf{87.3} \\
& $\phi_\text{F-FAemb}+\psi_\text{d}$ & 88.6 & \textbf{88.4} &  88.5 & 86.4\\
\hline
\end{tabular}
\vspace{-0.5em}
\end{table}
\subsubsection{Final feature dimensionality} 
\label{sssec:final-dim}

Since our framework provides the flexibility of choosing different dimensions for final representations, we evaluate the impact of final image representation on the retrieval performance.

\begin{wraptable}{r}{0.5\textwidth}
\vspace{-0.5em}
\caption{Number of retained PCA components (of local conv. features) and codebook size (of T-emb) for different dimensionalities.}
\label{tb:vary-dim}
\centering
\begin{tabular}{|l|c|c|c|c|c|}
\hline
\textbf{Dim.} $D$ & 512 & 1024 & 2048 & 4096 & 8064\\
\hline
$d$ & 32 & 64 & 64 & 64 & 128 \\
$|\mathcal{C}|$ & 20 & 18 & 34 & 66 & 64 \\
\hline
\end{tabular}
\end{wraptable}
Considering our default framework --- $\mathcal{M}_* + \phi_\Delta+\psi_\text{d}$, we empirically set the number of retained PCA components (of local conv. features) and the codebook size for different dimensionalities in Table \ref{tb:vary-dim}. For compact final representations of 512-D, we choose $d = 32$ to avoid using too few visual words as this drastically degrades performance \cite{Temb}. 
For longer final representations, i.e. 1024, 2048, 4096, imitating Fisher and  VLAD presentations for SIFT features~\cite{DBLP:journals/pami/JegouPDSPS12}, we reduce local conv. features to $d=64$. For the largest considered representation, i.e. 8064,  imitating the Temb representation for SIFT features \cite{Temb}, we reduce local conv. features to $d=128$. Note that the settings in Table \ref{tb:vary-dim} are applied for all later experiments.

\begin{wrapfigure}{r}{0.6\textwidth}
\small
\centering
\begin{subfigure}[b]{0.28\textwidth}
\includegraphics[width=\textwidth]{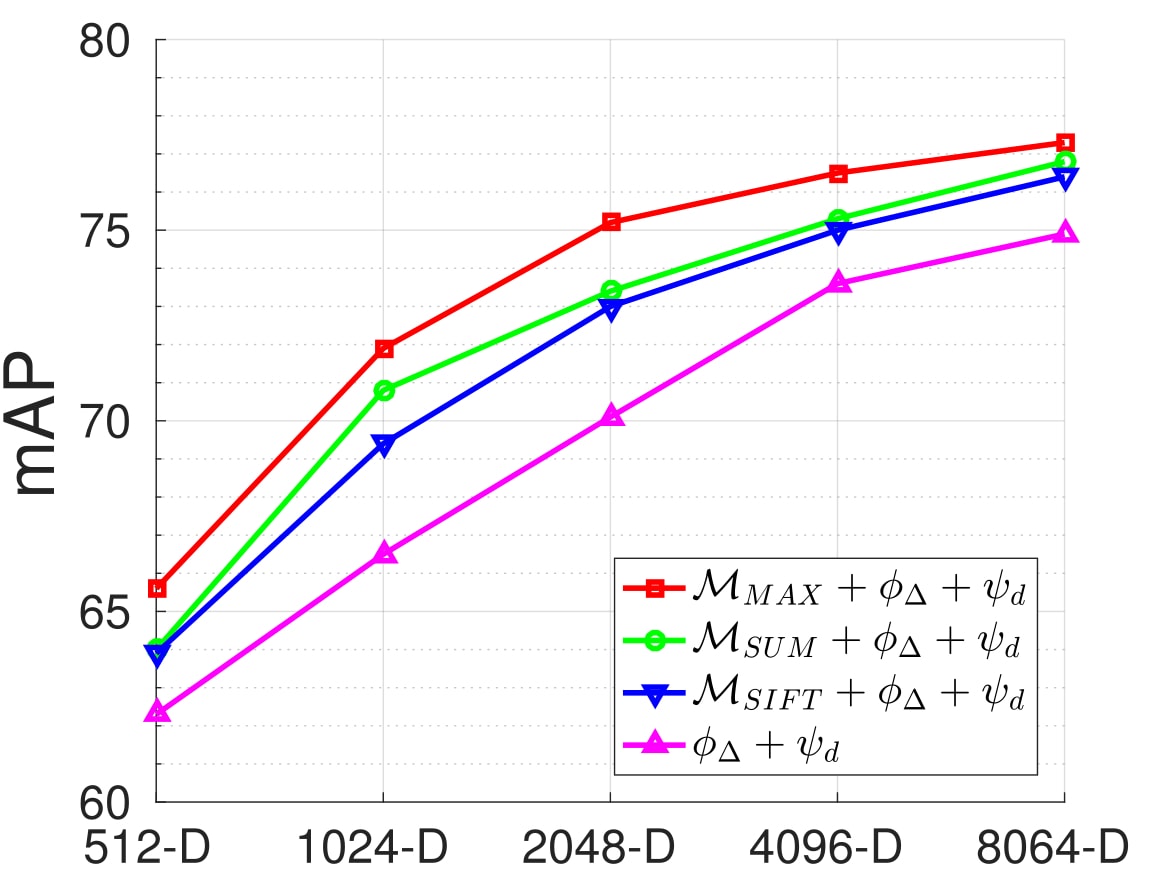}
\caption{\textbf{\textit{Oxford5k}}}
\end{subfigure}
\begin{subfigure}[b]{0.28\textwidth}
\includegraphics[width=\textwidth]{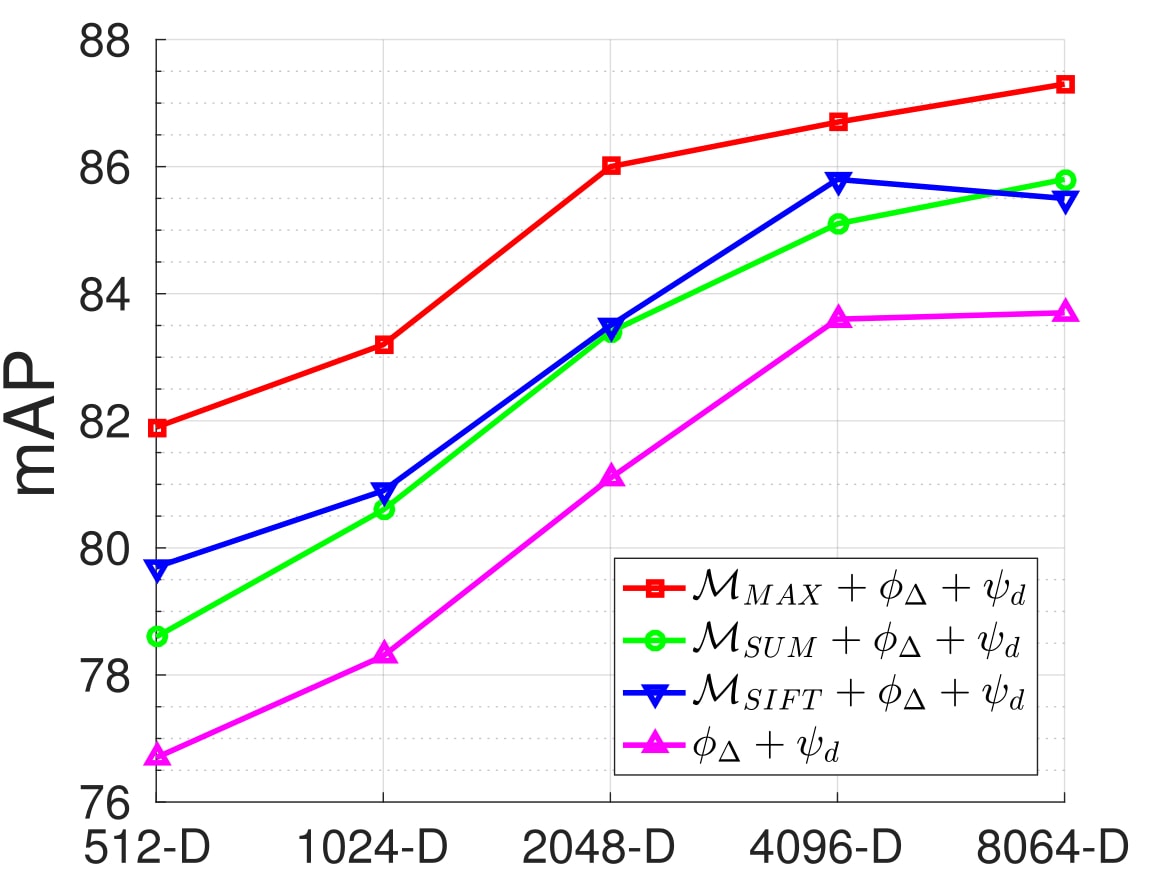}
\caption{\textbf{\textit{Paris6k}}}
\end{subfigure}
\caption{Impact of the final representation dimensionality on retrieval performance.
}
\label{fig:compare-mask-vary-D}
\vspace{-1em}
\end{wrapfigure}

The Figure \ref{fig:compare-mask-vary-D} shows the retrieval performances at different final feature dimensionalities for {\textit{Oxford5k}} and {\textit{Paris6k}} datasets. Unsurprisingly, the proposed framework can achieve higher performance gains when the final feature dimensionality increases. At 4096-D or higher, the improvements become small (or even decreased for $\mathcal{M}_{SIFT} + \phi_\Delta+\psi_\text{d}$ scheme on \textit{Paris6k} dataset). 
More important, the masking schemes consistently boost retrieval performances across different dimensionalities.

\subsubsection{Image size}
\label{ssec:img_size}

\begin{wraptable}{r}{0.6\textwidth}
\small
\caption{Impact of different input image sizes on retrieval performance.
The framework of $\mathcal{M}_\text{MAX/SUM} + \phi_\Delta + \psi_\text{d}$ is used to produce image representations.}
\label{tb:compare-image-size}
\centering
\begin{tabular}{|c|c|c|c|c|c|}
\hline
\multirow{2}{*}{\textbf{Dim.} $D$} & \multirow{2}{*}{$\max(W_I,H_I)$} & \multicolumn{2}{c|}{\textbf{Oxford5k}} &  \multicolumn{2}{c|}{\textbf{Paris6k}} \\
\cline{3-6} & & $\mathcal{M}_\text{SUM}$ & $\mathcal{M}_\text{MAX}$ & $\mathcal{M}_\text{SUM}$ & $\mathcal{M}_\text{MAX}$ \\
\hline
\multirow{2}{*}{512} & 724 & 56.4 & 60.9 & 79.3 & 81.2\\
                     & 1024 & 64.0 & 65.7 & 78.6 & 81.6  \\
\hline
\end{tabular}
\end{wraptable}

Since our framework highly depends on the number of local conv. features, it is necessary to evaluate the performance of our framework with a smaller image size. 	We present the retrieval performance on \textit{Oxford5k} and \textit{Paris6k} datasets with the image sizes of $\max(W_I,H_I)=1024$ and $\max(W_I,H_I)=724$ in Table \ref{tb:compare-image-size}. Similar to the reported results of \cite{R-MAC} on {\textit{Oxford5k}} dataset, we observe around 6-7\% drop in mAP when using smaller input images of $\max(W_I,H_I)=724$ rather than the original images. While on {\textit{Paris6k}} dataset, interestingly, the performances are more stable to changes of the image size. We observe a small drop of 2.2\% mAP on \textit{Paris6k} dataset for R-MAC \cite{R-MAC} with our experiments. The experimental results suggest that R-MAC \cite{R-MAC} and our methods are equivalently affected by the change in the image size.

The large performance drops on {\textit{Oxford5k}} can be explained that with higher resolution images, the CNN can take a closer \textit{``look''} on smaller details in the images. Hence, the local conv. features can better distinguish details in different images. 
While the stable performance on \textit{Paris6k} dataset can be perceived that the differences among scenes are at global structures, i.e., a higher abstract level, instead of small details as on \textit{Oxford5k} dataset. 
This explanation is actually consistent with human understanding on these datasets.


\subsubsection{Layer selection} 
\label{ssec:layer_sel}
\begin{wrapfigure}{tr}{0.6\textwidth}
\centering
\begin{subfigure}[b]{0.28\textwidth}
\includegraphics[width=\textwidth]{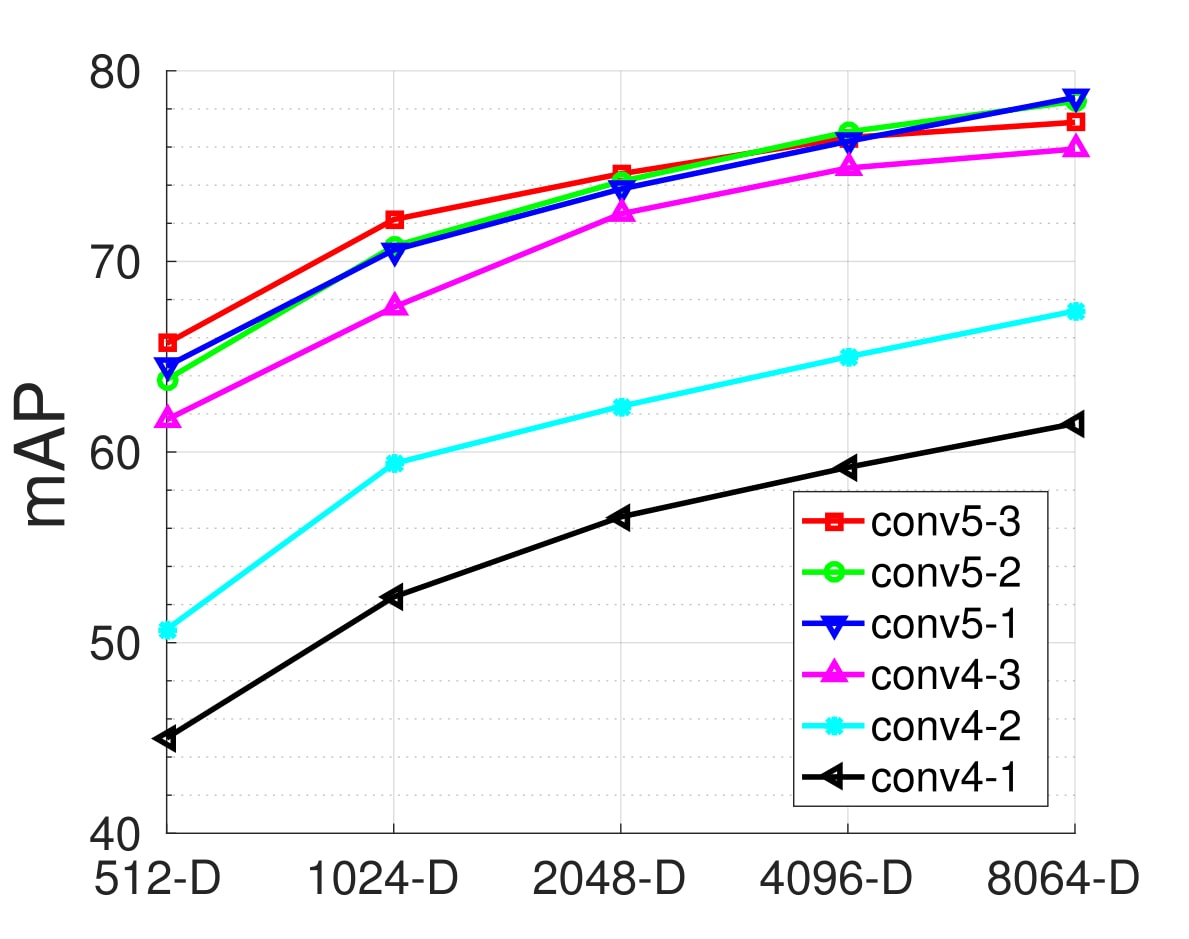}
\caption{\textbf{\textit{Oxford5k}}}
\end{subfigure}
\begin{subfigure}[b]{0.28\textwidth}
\includegraphics[width=\textwidth]{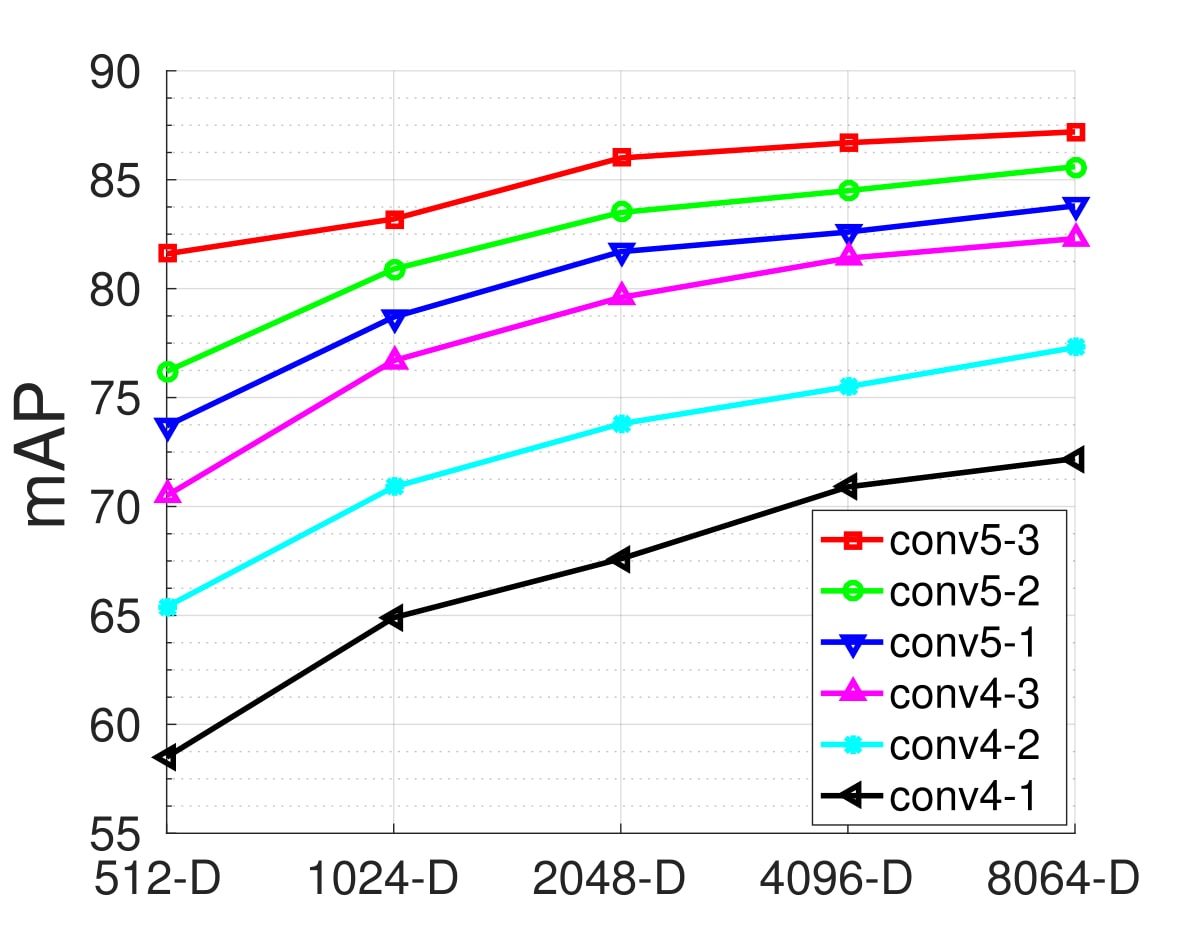}
\caption{\textbf{\textit{Paris6k}}}
\end{subfigure}
\caption{Impact of local deep conv. features from different layers on retrieval performance.
The framework of $\mathcal{M}_\text{MAX} + \phi_\Delta + \psi_\text{d}$ is used to produce image representations.}
\label{fig:compare-layers}
\vspace{-0.5em}
\end{wrapfigure}

In \cite{cnn_max_pooling}, 
the authors mentioned that deeper conv. layers produce features that are more reliable in differentiating images.
Here, we re-evaluate this statement using our proposed framework by comparing the retrieval performance (mAP) of features extracted from different conv. layers, including $\mathtt{conv5\text{-}3}$, $\mathtt{conv5\text{-}2}$, $\mathtt{conv5\text{-}1}$, $\mathtt{conv4\text{-}3}$, $\mathtt{conv4\text{-}2}$, and $\mathtt{conv4\text{-}1}$, at the same dimensionality. 
The experimental results on \textit{Oxford5k} and \textit{Paris6k} datasets are shown in  Figure \ref{fig:compare-layers}. We observe that the performances are slightly decreased when using lower conv. layers until $\mathtt{conv4\text{-}3}$. It means that conv. features of these layers, e.g., $\mathtt{conv5\text{-}3}$, $\mathtt{conv5\text{-}2}$, $\mathtt{conv5\text{-}1}$, $\mathtt{conv4\text{-}3}$, are still very discriminative. Hence, combining information of these layers may be beneficial. However, when going down further to $\mathtt{conv4\text{-}2}$ and $\mathtt{conv4\text{-}1}$, the performances are significantly lower. 
In summary, regarding the pre-trained VGG network \cite{VGG}, the last conv. layer (i.e., $\mathtt{conv5\text{-}3}$) produces the most reliable representation for image retrieval.

Furthermore, as assembling multiple conv. layers of CNN would be beneficial  \cite{hypercolumns} in localizing the saliency objects, we conduct additional experiments to evaluate whether combining different levels of abstraction from different conv. layers of CNN be useful for the retrieval task. Specifically, we concatenate feature maps from different layers as hyper-column feature maps which allow to normally use the proposed masking schemes. 
The experimental results are reported in \ref{tb:hypercolumn}, from which we observe that combining multiple conv. layers as hyper-column features helps to improve performances across many datasets, e.g., \textit{Oxford5k}, \textit{Paris6k}, and \textit{Holidays}.

\begin{table}[!t]
\footnotesize
\caption{Impact of combining multiple conv. layers as hyper-column feature maps on \textbf{\textit{Oxford5k}}, \textbf{\textit{Paris6k}}, and \textbf{\textit{Holidays}} datasets. The framework of $\mathcal{M}_\text{MAX} + \phi_{\Delta(64,18)} + \psi_\text{d}$ is used to produce image representations, where $\phi_{\Delta(64,18)}$ denotes Temb with $d=64$ and $|\mathcal{C}|=18$.}
\label{tb:hypercolumn}
\small
\centering
\begin{tabular}{|c|c|c|c c c|}
\hline
$\mathtt{conv5\text{-}3}$ & $\mathtt{conv5\text{-}2}$ & $\mathtt{conv5\text{-}1}$ & \textbf{Oxford5k} & \textbf{Paris6k} & \textbf{Holidays} \\
\hline
\checkmark &  &  & 72.2 & 83.2 & 88.4 \\
\checkmark & \checkmark &  & 73.3 & 83.5 & 90.4 \\
\checkmark &  & \checkmark & 74.2 & 83.8 & 90.9 \\
\checkmark & \checkmark & \checkmark & 74.8 & 84.5 & 90.8 \\
\hline
\end{tabular}
\vspace{-0.5em}
\end{table}

\subsubsection{Binary representation framework}
\label{subsub:bin}
\begin{table*}[t]
\small
\caption{Comparison of different frameworks when the final representations are binary values. The values in brackets in \textbf{Embedding} column indicate the dimension of local conv. features after PCA and the codebook size, respectively. \textbf{Dim.} column indicates the dimension of real-valued representations before subjecting into a hash function. We evaluate the binary representations at code lengths $64, 128, 256, 512$ with three state-of-the-art unsupervised hashing methods ITQ~\cite{ITQ}, RBA~\cite{SAH}, and KMH~\cite{kmeanHash}. Results are reported on \textbf{\textit{Oxford5k}}, \textbf{\textit{Paris6k}} and \textbf{\textit{Holidays}} dataset.
}
\label{tb:compare_hash_framework}
\centering
\begin{tabular}{|c|c|c|c c c c|c c c c|c c c c|}
\hline
\multirow{2}{*}{} & \multirow{2}{*}{\textbf{Embed}} & \multirow{2}{*}{\textbf{Dim.}} &  \multicolumn{4}{c|}{\textbf{Oxford5k}} & \multicolumn{4}{c|}{\textbf{Paris6k}} & \multicolumn{4}{c|}{\textbf{Holidays}} \\
\cline{4-15}
& & & 64 & 128 & 256 & 512 & 64 & 128 & 256 & 512 & 64 & 128 & 256 & 512 \\
\hline\hline
\multirow{4}{*}{\rotatebox[origin=c]{90}{ITQ \cite{ITQ}}} & $\phi_{\Delta(32,20)}$ & 512 & 18.3 & \textbf{31.6} & 45.0 & \textbf{57.3} & \textbf{32.9} & 49.7 & 63.0 & 74.7 & {57.5} & 70.7 & \textbf{79.5} & \textbf{83.5} \\
& $\phi_{\Delta(64,18)}$ & 1024 & \textbf{18.7} & 29.5 & 42.9 & 55.8 & 33.5 & 49.0 & 61.0 & 72.4 & \textbf{58.7} & 71.5 & 79.4 & 82.9 \\
& $\phi_{\Delta(64,34)}$ & 2048 & 18.3 & 27.7 & 38.4 & 50.3 & 28.4 & 44.3 & 57.9 & 68.6 & 57.9 & 71.1 & 79.1 & 82.3 \\
& $\phi_{\Delta(64,66)}$ & 4096 & 16.0 & 23.0 & 33.6 & 45.8 & 26.1 & 40.1 & 52.9 & 65.4 & 56.1 & 70.2 & 78.8 & 80.5 \\
\hline

\multirow{4}{*}{\rotatebox[origin=c]{90}{RBA \cite{SAH}}} & $\phi_{\Delta(32,20)}$  & 512 & 17.8 & 31.3 & \textbf{45.3} & 57.1 & 31.5 & \textbf{50.8} & \textbf{63.7} & \textbf{74.9} & 56.7 & \textbf{71.6} & 78.5 & 83.2 \\
& $\phi_{\Delta(64,18)}$ & 1024 & 18.4 & 30.1 & 42.7 & 55.7 & 32.8 & 49.4 & 61.3 & 72.8 & 57.6 & 71.1 & 79.3 & 82.7 \\
& $\phi_{\Delta(64,34)}$ & 2048 & 17.3 & 30.9 & 39.1 & 53.5 & 29.0 & 45.0 & 59.1 & 68.6 & 57.1 & 70.8 & 78.8 & 81.8 \\
& $\phi_{\Delta(64,66)}$ & 4096 & 15.7 & 25.1 & 39.0 & 48.3 & 25.8 & 39.3 & 55.6 & 64.6 & 55.6 & 67.5 & 77.5 & 80.1 \\
\hline
\multirow{4}{*}{\rotatebox[origin=c]{90}{KMH \cite{kmeanHash}}} &  $\phi_{\Delta(32,20)}$  & 512 & 18.5 & 26.5 & 39.1 & 54.4 & 32.0 & 45.7 & 61.6 & 75.3 & 53.4 & 65.0 & 75.0 & 80.8 \\
& $\phi_{\Delta(64,18)}$ & 1024 & 18.1 & 28.1 & 41.4 & 53.5 & 30.0 & 48.3 & 61.4 & 73.7 & 54.6 & 68.0 & 78.0 & 82.4 \\
& $\phi_{\Delta(64,34)}$ & 2048 & 15.7 & 26.7 & 38.5 & 51.4 & 26.0 & 40.8 & 56.7 & 69.2 & 51.2 & 68.4 & 76.4 & 81.2 \\
& $\phi_{\Delta(64,66)}$ & 4096 & 13.4 & 20.7 & 31.2 & 47.0 & 21.1 & 33.3 & 49.8 & 63.7 & 50.0 & 63.2 & 72.8 & 80.1 \\
\hline

\end{tabular}
\end{table*}

In this section, we conduct experiments at a wide range of settings to find the setting that  produces the best binary representation. As discussed in section \ref{ssec:img_size} and \ref{ssec:layer_sel}, when using images of $\max(W_I, H_I)=1024$, the last conv. layer of the VGG network \cite{VGG}, i.e., $\mathtt{conv5\text{-}3}$, produces the most reliable representations. 
Hence, the default framework of $\mathcal{M}_\text{MAX} + \phi_\Delta + \psi_\text{d}$, in combination with these two settings (i.e., $\max(W_I, H_I)=1024$ and  $\mathtt{conv5\text{-}3}$ features), is used to produce real-valued representations before passing forward to a hashing module.
Furthermore, in the literature, unsupervised hashing methods are usually proposed to work with hand-crafted global image features, e.g., GIST \cite{gist}, or deep learning features of a fully-connected layer, e.g., $\mathtt{fc7}$ of AlexNet or VGG, it is unclear which method works the best with our proposed aggregated representations. Hence, we conduct experiments with various state-of-the-art unsupervised hashing methods including iterative quantization (ITQ)~\cite{ITQ}, relaxed binary autoencoder (RBA)~\cite{SAH}, and K-means hashing (KMH)~\cite{kmeanHash} to find the best hashing module for our framework.

The experimental results on \textit{Oxford5k}, \textit{Paris6k}, and \textit{Holidays} datasets are presented in Table \ref{tb:compare_hash_framework}. There are some main observations from the results. Firstly, at the same code length, ITQ and RBA achieve comparable results, while both these methods are significantly outperforms KHM, on all datasets. Secondly, as discussed in Section \ref{sssec:final-dim}, embedding local conv. features to a higher dimensional space helps to enhance the discriminative power of the real-valued aggregated representation; however, as shown in Table~\ref{tb:compare_hash_framework}, embedding to too high-dimensional space also causes information loss when producing compact binary codes, i.e., the best mAPs are achieved when the aggregated representation are at 512 or 1024 dimensions. The higher dimensional representations (i.e., 2048-D or 4906-D) cause the more mAPs loss. As embedding to 512-D, i.e., $\phi_{\Delta(32,20)}$, gives most stable results, we use this configuration in our final framework when producing binary representations.


\begin{table*}[t]
\caption{Comparison with the state of the art when the final representations are real values. The results of compared methods are cited from the corresponding papers when available. For results of R-MAC\cite{R-MAC} on Holidays and Holidays+Flickr1M, we use the released code of R-MAC\cite{R-MAC} to evaluate on these datasets.}

\label{tb:state-of-art}
\footnotesize
\centering
\begin{tabular}{|c|l|c|c|c|c|c|c|c|c|}
\hline
 & \multirow{2}{*}{\textbf{Methods}} & \multirow{2}{*}{\textbf{Dim.}} & \textbf{Size} &\multicolumn{6}{c|}{\textbf{Datasets}}\\
\cline{5-10} & & & \textbf{(Byte)} & \textit{\textbf{Oxf5k}} & \textbf{\textit{Oxf105k}} & \textbf{\textit{Par6k }}& \textbf{\textit{Par106k}} & \textbf{\textit{Hol}} & \textbf{\textit{Hol+Fl1M}} \\
\hline
\hline

\multirow{4}{*}{\rotatebox[origin=c]{90}{\textbf{SIFT}}}  & $\phi_\Delta+\psi_\text{d}$ \cite{Temb} & 512 & 2k & 52.8 & 46.1 & - & - & 61.7 & 46.9\\ 
& $\phi_\Delta+\psi_\text{d}$ \cite{Temb} & 1024 & 4k & 56.0 & 50.2 & - & - & 72.0 & 49.4 \\
& $\phi_\text{F-FAemb}+\psi_\text{d}$ \cite{F-FAemb} & 512 & 2k & 53.9 & 50.9 & - & - & 69.0 & 65.3 \\
& $\phi_\text{F-FAemb}+\psi_\text{d}$ \cite{F-FAemb} & 1024 & 4k & 58.2 & 53.2 & - & - & 70.8 & 68.5 \\
\hline\hline
\multirow{14}{*}{\rotatebox[origin=c]{90}{\textbf{Off-the-shelf network}}} & SPoC \cite{cnn_max_pooling} & 256 & 1k & 53.1 & - & 50.1 & - & 80.2 & - \\
& MOP-CNN \cite{MOP} & 512 & 2k & - & - & - & - & 78.4 & - \\
& CroW \cite{CroW} & 512 & 2k & 70.8 & 65.3 & 79.7 & 72.2 & 85.1 & - \\
& MAC \cite{finetune_hard_samples} & 512 & 2k & 56.4 & 47.8 & 72.3 & 58.0 & 76.7 & - \\
& R-MAC \cite{R-MAC} & 512 & 2k & 66.9 & 61.6 & 83.0 & 75.7 & 86.6 & 71.5 \\
& NetVLAD \cite{netvlad} & 1024 & 4k & 62.6 & - & 73.3 & - & 87.3 & - \\
& PWA \cite{PWA} & 1024 & 4k & \textbf{75.3} & 69.3 & 84.2 & 78.2 & - & - \\
& NetVLAD \cite{netvlad} & 4096 & 16k & 66.6 & - & 77.4 & - & 88.3 & - \\

\cline{2-10}
& \textbf{$\mathcal{M}_\text{SIFT} + \phi_\Delta+\psi_\text{d}$} & 512 & 2k & 64.4 & 59.4 & 79.5 & 70.6 & 86.5 & - \\
& \textbf{$\mathcal{M}_\text{SUM} + \phi_\Delta+\psi_\text{d}$} & 512 & 2k &  64.0 & 58.8 & 78.6 & 70.4 & 86.4 & - \\
& \textbf{$\mathcal{M}_\text{MAX} + \phi_\Delta+\psi_\text{d}$} & 512 & 2k &  65.7 & 60.5 & 81.6 & 72.4 & 85.0 & 71.9 \\
& \textbf{$\mathcal{M}_{\text{MAX}(\mathtt{conv5\text{-}3,2,1})} + \phi_\Delta+\psi_\text{d}$} & 512 & 2k &  69.2 & 65.3 & 82.5 & 74.0 & 88.7 & 73.0 \\

& \textbf{$\mathcal{M}_\text{SIFT} + \phi_\Delta+\psi_\text{d}$} & 1024 & 4k & 69.9 & 64.3 & 81.7 & 73.8 & 87.1 & - \\
& \textbf{$\mathcal{M}_\text{SUM} + \phi_\Delta+\psi_\text{d}$} & 1024 & 4k &  70.8 & 64.4 & 80.6 & 73.8 & 86.9 & -\\
& \textbf{$\mathcal{M}_\text{MAX} + \phi_\Delta+\psi_\text{d}$} & 1024 & 4k &  72.2 & 67.9 & 83.2 & 76.1 & 88.4 & 79.1 \\
& \textbf{$\mathcal{M}_{\text{MAX}(\mathtt{conv5\text{-}3,2,1})} + \phi_\Delta+\psi_\text{d}$} & 1024 & 4k &  {74.8} & \textbf{70.4} & \textbf{84.5} & \textbf{78.6} & \textbf{90.8} & \textbf{81.7}  \\
\hline
\hline

\multirow{9}{*}{\rotatebox[origin=c]{90}{\textbf{Finetuned network}}} & siaMAC $+$ MAC \cite{finetune_hard_samples} & 512 & 2k & 79.7 & 73.9 & 82.4 & 74.6 & 79.5 & - \\
& siaMAC $+$ R-MAC \cite{finetune_hard_samples} & 512 & 2k & 77.0 & 69.2 & 83.8 & 76.4 & 82.5 & - \\
& NetVLAD$\star$ \cite{netvlad} & 1024 & 4k & 69.2 & - & 76.5 & - & 86.5 & -  \\
&  NetVLAD$\star$ \cite{netvlad} & 4096 & 16k & 71.6 & - & 79.7 & - & 87.5 & -  \\
\cline{2-10}
& siaMAC\textbf{$ + \mathcal{M}_\text{MAX} + \phi_\Delta+\psi_\text{d}$} & 512 & 2k &  77.7 & 72.7 & 83.2 & 76.5 & 86.3 & -  \\
& siaMAC \textbf{$ +\mathcal{M}_\text{MAX} + \phi_\Delta+\psi_\text{d}$} & 1024 & 4k &  81.4 & 77.4 & 84.8 & 78.9 & 88.9 & \textbf{82.1} \\
& NetVLAD$\star$  \textbf{$ +\mathcal{M}_\text{MAX} + \phi_\Delta+\psi_\text{d}$} & 1024 & 4k & 75.2 & 71.7 & 84.4 & 76.9 & 91.5 & - \\
\cline{2-10}
& NetVLAD$\star$ \textbf{$ +\mathcal{M}_\text{MAX} + \phi_\Delta+\psi_\text{d}$} & 4096 & 16k & 78.2 & 75.7 & 87.8 & 81.8 & \textbf{92.2} & -  \\
& siaMAC  \textbf{$+\mathcal{M}_\text{MAX} + \phi_\Delta+\psi_\text{d}$} & 4096 & 16k & \textbf{83.8} & \textbf{80.6} & \textbf{88.3} & \textbf{83.1} & 90.1 & -  \\
\hline
\end{tabular}
\vspace{-0.5em}
\end{table*}

\subsection{Comparison to the state of the art}
\label{sssec:compare_sota}
We comprehensively evaluate and compare our proposed framework with the state of the art in the image retrieval task. We separate two experimental settings. The first experiment is when images are represented by mid-dimensional real-valued presentations. The second experiment is when images are represented by very compact representations, i.e., very short real-valued vectors or binary vectors. 

\subsubsection{Comparison with the state of the art when images are represented by mid-dimensional real-valued vectors}
\label{sssec:real_value_compare}
We report comparative results when images are represented by real-valued vectors in Table \ref{tb:state-of-art}. We separate two different settings, i.e., when deep features are extracted from an off-the-shelf pretrained VGG network and are extracted from a VGG network which is fine-tuned for the image retrieval task. 

\smallskip
\textbf{Using off-the-shelf VGG network \cite{VGG}.} The first observation is that at the dimensionality of 1024, our framework using MAX-mask ($\mathcal{M}_\text{MAX} + \phi_\Delta+\psi_\text{d}$) achieves the best mAP in comparison to recent deep learning-based methods \cite{cnn_max_pooling,CroW,finetune_hard_samples,R-MAC,netvlad} acrossing different datasets.  The second observation is that by combining multiple conv. layers, $\mathtt{conv5\text{-}3}$, $\mathtt{conv5\text{-}2}$, and  $\mathtt{conv5\text{-}1}$, denoted as $\mathcal{M}_{\text{MAX}(\mathtt{conv5\text{-}3,2,1})}$, the proposed MAX scheme consistently boosts the retrieval accuracy. 
Our framework (\textbf{$\mathcal{M}_{\text{MAX}(\mathtt{conv5\text{-}3,2,1})} + \phi_\Delta+\psi_\text{d}$}) with dimensionality of 512 is competitive with other state-of-the-art methods \cite{CroW,R-MAC}. In particular, in comparison with CroW \cite{CroW}, while having slightly lower performances in \textit{Oxford5k}, our method outperforms CroW on \textit{Paris6k} and \textit{Holidays}. 
In comparison with R-MAC \cite{R-MAC}, the proposed framework outperforms RMAC on  \textit{Oxford5k} and \textit{Holidays} datasets, while it is comparable to RMAC on \textit{Paris6k} dataset. Note that in some comparison methods, e.g., siaMAC~\cite{finetune_hard_samples}, R-MAC~\cite{R-MAC}, SPoC \cite{cnn_max_pooling}, CroW~\cite{CroW}, the dimensionality is 256 or 512. This is due to the final representation dimensionality of these methods is upper bounded by the number of feature channels $K$ of the selected network architecture and layers, e.g., $K=512$ for $\mathtt{conv5}$ of VGG16. Our proposed method, on the other hand, provides more flexibility in the representation dimensionality, thanks to the embedding process.

It is worth noting that NetVLAD \cite{netvlad}, MOP-CNN \cite{MOP}, and PWA \cite{PWA} methods can also produce higher dimensional representation by increasing the codebook size. However, as shown in Table~\ref{tb:state-of-art} at comparable dimensions, the proposed framework clearly outperforms NetVLAD and MOP-CNN. In addition, at the dimensionality of 1024, our framework $\mathcal{M}_{\text{MAX}(\mathtt{conv5\text{-}3,2,1})} + \phi_\Delta+\psi_\text{d}$ is slightly more favorable than PWA. Our framework achieves better performances on \textit{Oxford105k}, \textit{Paris6k}, and \textit{Paris106k} datasets and is only slightly lower in mAP in \textit{Oxford5k} dataset.  

\smallskip
\textbf{Taking advantages of fine-tuned VGG networks.} Since our proposed framework takes the 3D activation tensor of a conv. layer as the input, it is totally compatible with deep networks which are fine-tuned for the image retrieval task such as siaMAC~\cite{finetune_hard_samples} and NetVLAD~\cite{netvlad} (noted as NetVLAD$\star$).  In the \textbf{``Fine-tuned network"} section of Table \ref{tb:state-of-art}, we evaluate our best framework --- $\mathcal{M}_\text{MAX} + \phi_\Delta+\psi_\text{d}$ in which the local conv. features of the fine-tuned VGG networks NetVLAD$\star$~\cite{netvlad}, siaMAC~\cite{finetune_hard_samples} are used as inputs. 

The experimental results from Table \ref{tb:state-of-art} show that, for our $\mathcal{M}_\text{MAX} + \phi_\Delta+\psi_\text{d}$ framework,  using conv. features from the siaMAC network usually gives better performance than using those from the fine-tuned NetVLAD$\star$ network. 
When using local conv. features extracted from the fine-tuned siaMAC network \cite{finetune_hard_samples}, our method is competitive to \textbf{siaMAC $+$ R-MAC} and \textbf{siaMAC $+$ MAC}~\cite{finetune_hard_samples} at dimensionality of 512. 
At 1024 dimensions, our method consistently outperforms both siaMAC and NetVLAD$\star$ on all datasets. 
These considerable improvements indicate that our proposed framework can fully utilize the discrimination gain of local conv. features achieved by those fine-tuning networks.

\smallskip
\textbf{Very large scale image retrieval.} In order to verify the capabilities of our framework in real scenarios, we now evaluate it with a very large scale dataset, \textit{Holidays$+$Flickr1M}. The experimental results show that the proposed framework ``siaMAC + $\mathcal{M}_\text{MAX} + \phi_\Delta+\psi_\text{d}$" is quite robust to the database size, i.e., when adding 1M distractor images to the Holidays dataset, the performance drop is only about 7\%. We achieve a mAP of  82.1 which is significantly higher than 71.5 of R-MAC \cite{R-MAC}. 

\smallskip
\begin{wraptable}{r}{0.6\textwidth}
\vspace{-0.5em}
\footnotesize
\caption{Comparison with SCDA \cite{SCDA}}
\label{tb:compare_scda}
\centering
\begin{tabular}{|l|c|c|c|c|c|c|c|c|}
\hline
\multirow{2}{*}{\textbf{Method}} & \multirow{2}{*}{\textbf{Dim.}} & \multicolumn{2}{c|}{\textbf{\textit{mAP}}} \\
\cline{3-4}
& & \textbf{\textit{Oxford5k}} & \textbf{\textit{Holidays}}\\
\hline
SCDA\cite{SCDA} & 4096 $\to$ 512 & 67.7 & 92.1 \\
\textbf{$\mathcal{M}_\text{SUM} + \phi_{\Delta(64,64)}+\psi_\text{d}$} & 4096 $\to$ 512 & 77.2 & 92.0 \\
\textbf{$\mathcal{M}_\text{MAX} + \phi_{\Delta(64,64)}+\psi_\text{d}$} & 4096 $\to$ 512 & 78.6 & 93.2 \\
\hline
\end{tabular}
\vspace{-0.5em}
\end{wraptable}


\textbf{Comparison to Selective Convolutional Descriptor Aggregation (SCDA) \cite{SCDA}.} Recently, Wei et al. \cite{SCDA} proposed a method which  selects deep features on $relu5\_2$ and $pool\_5$ layers from the pretrained VGG networks. Their  method shares some similarities with our SUM-mask. Our work, however, is different from \cite{SCDA} in several improtant aspects, i.e., we propose and evalute various masking schemes, i.e., SUM-mask, SIFT-mask, and MAX-mask. Our experiments show that the MAX-mask scheme consitently outperforms other schemes. In addition, we utilize state-of-the-art embedding and aggregating methods to enhance the discirminative power of the final representation.  In order to have a complete evaluation to SCDA~\cite{SCDA}, we conduct comparison with SCDA \cite{SCDA} on \textit{Oxford5k} and \textit{Holidays} datasets. We exactly follow the setting of SCDA, i.e., the 5063 and 1491 gallery images of \textit{Oxford5k} and \textit{Holidays} datasets are used as the training set when learning codebooks for the embedding. In this experiment, for our methods, we do not truncate the first low frequency components when embedding. This makes the original dimension of the aggregated representations of SCDA and our methods are comparable, i.e., 4096. The low dimensionality, i.e. 512, is achieved by applying PCA.  
The comparative results in Table \ref{tb:compare_scda} clearly show the superior performances, especially on \textit{Oxford5k} dataset, of our proposed framework over SCDA.

\begin{table*}[!t]
\caption{Comparison with the state of the art on very compact representations. (re.) and (bin.) mean that the representations are real values and binary values, respectively. For real values, the distance meansure is cosine and for binary values, distance meansure is Hamming.
}
\label{tb:state-of-art_compact}
\footnotesize
\centering
\begin{tabular}{|l|c|c|c|c|c|c|c|}
\hline
 \multirow{2}{*}{\textbf{Method}} & \multirow{2}{*}{\textbf{Dim.}} & \textbf{Size} & \multicolumn{5}{c|}{\textbf{Datasets}}\\
\cline{4-8} & & \textbf{(Byte)} & \textit{\textbf{Oxf5k}} & \textbf{\textit{Oxf105k}} & \textbf{\textit{Par6k }}& \textbf{\textit{Par106k}} & \textbf{\textit{Hol}} \\
\hline
\hline
siaMAC $+$ MAC \cite{finetune_hard_samples} & 16 (re.) & 64 & 56.2 & 45.5 & 57.3 & 43.4 & 51.3 \\
siaMAC $+$ R-MAC \cite{finetune_hard_samples} & 16 (re.)& 64 & 46.9 & 37.9 & 58.8 & 45.6 & 54.4 \\
GeM \cite{ft_nohuman} & 16 (re.) & 64 & 56.2 & 44.4 & 63.5 & 45.5 & 60.9 \\
\hline
\textbf{$\mathcal{M}_\text{MAX} + \phi_{\Delta(32,20)}+\psi_\text{d}+\text{ITQ}$} & {256} (bin.) & 32 & 45.0 & 38.3 & 63.0 & 50.5 & 79.5 \\
siaMAC \textbf{$+\mathcal{M}_\text{MAX} + \phi_{\Delta(32,20)}+\psi_\text{d}+\text{ITQ}$} & {256} (bin.) & 32 & {58.5} & {49.1} & {74.1} & {63.6} & {79.9} \\
\hline
\textbf{$\mathcal{M}_\text{MAX} + \phi_{\Delta(32,20)}+\psi_\text{d}+\text{ITQ}$} & {512} (bin.) & 64 & 57.3 & 49.8 & 74.7 & 56.1 & 83.5  \\
siaMAC \textbf{$+\mathcal{M}_\text{MAX} + \phi_{\Delta(32,20)}+\psi_\text{d}+\text{ITQ}$} & {512} (bin.) & 64 & \textbf{68.9} & \textbf{60.9} & \textbf{79.1} & \textbf{70.3} & \textbf{83.6} \\
\hline
\end{tabular}
\vspace{-0.5em}
\end{table*}

\subsubsection{Comparison with the state of the art when images are represented by very compact representations}
\label{sssec:compare_hash}
We now compare the quality of binary image representations producing by our framework with compact real-valued representations from state-of-the-art methods at comparable sizes (in Bytes) \cite{finetune_hard_samples,ft_nohuman}.
Furthermore, ITQ~\cite{ITQ} is used as the hashing function in our final framework since it gives competitive results (Section~\ref{subsub:bin}) and it is also computationally efficient in both training and producing new binary codes. We report the comparative results in Table \ref{tb:state-of-art_compact}. 

Firstly, we can observe that at the same image descriptor size, e.g., 64 bytes, even when using off-the-shelf VGG \cite{VGG}, our framework significantly outperforms \cite{finetune_hard_samples,ft_nohuman} which use fine-tuned VGG networks.  For examples, the proposed framwork outperforms the second best GeM~\cite{ft_nohuman} large margins, i.e., $+11.2\%$ and $+22.6\%$ on \textit{Paris6k} and \textit{Holidays} datasets, respectively. 
Secondly, when using local conv. features of a fine-tuned VGG, e.g., siaMAC \cite{finetune_hard_samples}, our framework achieves significant extra improvements in retrieval performances over all datasets. 


\vspace{-0.7em}
\subsection{Online processing time}
\label{ssec:processing_time}

\begin{wrapfigure}{tr}{0.40\textwidth}
\vspace{-1em}
\centering
\includegraphics[scale=0.38]{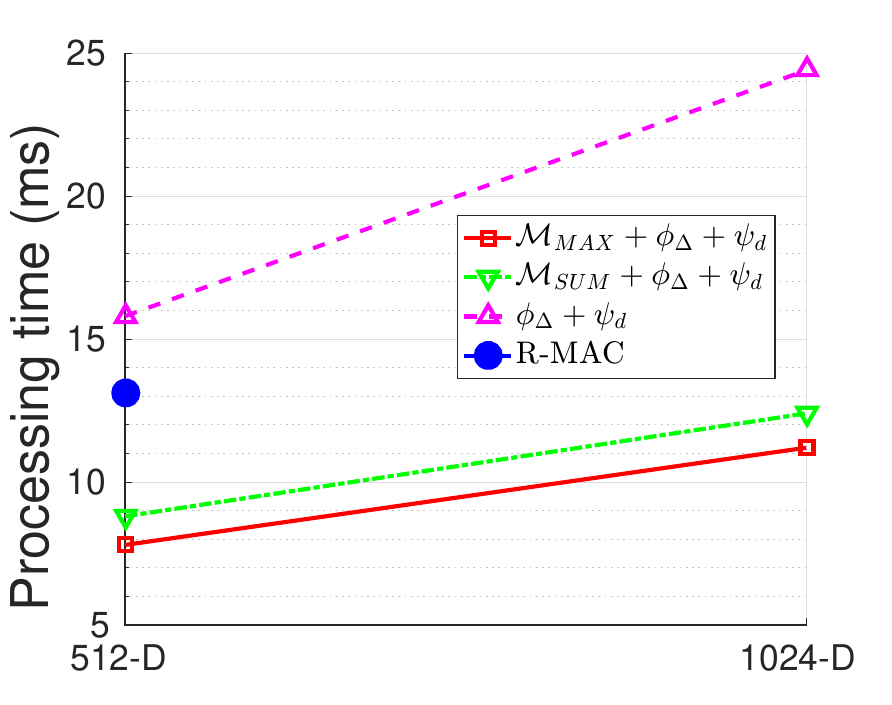}
\caption{The averaged online processing time images in \textit{Oxford5k} dataset.}
\label{fig:processing-time}
\vspace{-1em}
\end{wrapfigure}

We conduct experiments to empirically measure the online processing time of our proposed framework. We additionally compare our online processing time with one of the most competitive methods: 
R-MAC~\cite{R-MAC}. Both implementations of our framework and R-MAC are in Matlab. The experiments are executed on a workstation with a processor core (i7-6700 CPU @ 3.40GHz).  Fig. \ref{fig:processing-time} reports the averaged online processing time of \textit{Oxford5k} dataset ($5063$ images). Note that the processing time includes the time to compute and apply masks and excludes the time for extracting 3D convolutional feature maps. 
The figure clearly shows that the MAX/SUM-mask can help to considerably reduce the computational cost of our proposed framework. As the proposed masking schemes can eliminate about 70\% (for MAX-mask) and 50\% (for SUM-mask) of local conv. features (Section \ref{ssec:effectiveness}). Furthermore, at 512-D, our framework $\mathcal{M}_\text{MAX/SUM} + \phi_\Delta+\psi_\text{d}$ is faster than R-MAC \cite{R-MAC}.




\section{Conclusion}
\label{sec:conclusion}
In this paper, we present a novel, optimized and computationally-efficient framework for image retrieval task. The framework takes activations of a convolutional layer as the input and outputs a highly-discriminative image representation. In the framework, we propose to enhance discriminative power of the image representation in two main steps: (i) applying our proposed masking schemes, e.g., SIFT/SUM/MAX-mask, to select a subset of selective local conv. features, then (ii) employing the state-of-art embedding and aggregating methods \cite{Temb,F-FAemb}. In order to make the final representations suitable for large scale search, we further compress the real-valued representation by cascading a hashing function into framework. We comprehensively evaluate and analyze each component in the framework to figure out the best configuration. Solid experimental results show that our proposed framework favorably compares with the state of the art for real-valued representations. Moreover, our binary representations are significantly outperforms the state-of-the-art methods at comparable sizes.

\bibliographystyle{ACM-Reference-Format}
\bibliography{sigprocref}
\end{document}